\documentclass[sigconf]{acmart}

\usepackage{multirow}
\usepackage[font=footnotesize]{subfig}

\AtBeginDocument{%
  }

\setcopyright{acmcopyright}
\copyrightyear{2023}
\acmYear{2023}
\acmDOI{10.1145/3581783.3612471}

\acmConference[MM '23]{Proceedings of the 31st ACM International Conference on Multimedia}{October 29--November 3, 2023}{Ottawa, ON, Canada.}
\acmBooktitle{Proceedings of the 31st ACM International Conference on Multimedia (MM '23), October 29--November 3, 2023, Ottawa, ON, Canada}
\acmPrice{15.00}
\acmISBN{979-8-4007-0108-5/23/10}

\begin{document}

\title{SepMark: Deep Separable Watermarking for Unified Source Tracing and Deepfake Detection}

\author{Xiaoshuai Wu}
\email{shinewu@hnu.edu.cn}
\orcid{0000-0001-6241-3546}
\affiliation{%
	\department{College of Computer Science and Electronic Engineering}
	\institution{Hunan University, Changsha, China}
	\country{}
}

\author{Xin Liao}
\authornote{Corresponding author.}
\email{xinliao@hnu.edu.cn}
\orcid{0000-0002-9131-0578}
\affiliation{%
	\department{College of Computer Science and Electronic Engineering}
	\institution{Hunan University, Changsha, China}
	\country{}
}

\author{Bo Ou}
\email{oubo@hnu.edu.cn}
\orcid{0000-0001-6936-9955}
\affiliation{%
	\department{College of Computer Science and Electronic Engineering}
	\institution{Hunan University, Changsha, China}
	\country{}
}

\renewcommand{\shortauthors}{Xiaoshuai Wu et al.}

\begin{abstract}
Malicious Deepfakes have led to a sharp conflict over distinguishing between genuine and forged faces. Although many countermeasures have been developed to detect Deepfakes ex-post, undoubtedly, passive forensics has not considered any preventive measures for the pristine face before foreseeable manipulations. To complete this forensics ecosystem, we thus put forward the proactive solution dubbed SepMark, which provides a unified framework for source tracing and Deepfake detection. SepMark originates from encoder-decoder-based deep watermarking but with two separable decoders. For the first time the deep separable watermarking, SepMark brings a new paradigm to the established study of deep watermarking, where a single encoder embeds one watermark elegantly, while two decoders can extract the watermark separately at different levels of robustness. The robust decoder termed Tracer that resists various distortions may have an overly high level of robustness, allowing the watermark to survive both before and after Deepfake. The semi-robust one termed Detector is selectively sensitive to malicious distortions, making the watermark disappear after Deepfake. Only SepMark comprising of Tracer and Detector can reliably trace the trusted source of the marked face and detect whether it has been altered since being marked; neither of the two alone can achieve this. Extensive experiments demonstrate the effectiveness of the proposed SepMark on typical Deepfakes, including face swapping, expression reenactment, and attribute editing. Code will be available at \textcolor{blue}{\url{https://github.com/sh1newu/SepMark}}.
\end{abstract}

\begin{CCSXML}
<ccs2012>
<concept>
<concept_id>10002978.10002991.10002996</concept_id>
<concept_desc>Security and privacy~Digital rights management</concept_desc>
<concept_significance>500</concept_significance>
</concept>
</ccs2012>
\end{CCSXML}

\ccsdesc[500]{Security and privacy~Digital rights management}

\keywords{deep watermarking, deepfake forensics, watermarking robustness}

%% \received{20 February 2007}
%% \received[revised]{12 March 2009}
%% \received[accepted]{5 June 2009}

\maketitle

\section{Introduction}
\begin{figure}[t]
	\centering
	\includegraphics[width=\linewidth]{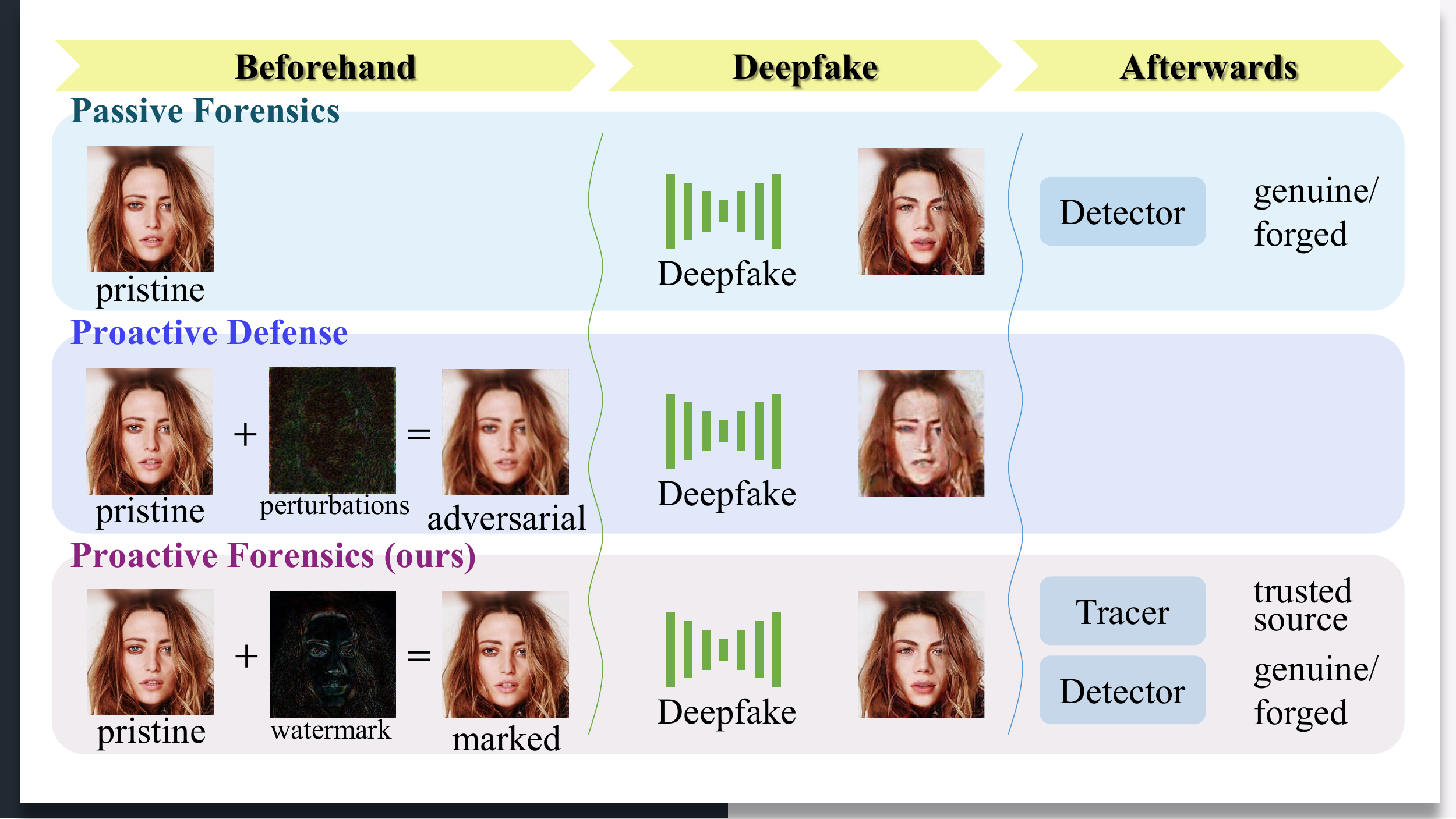}
	\caption{Illustration of different types of Deepfake countermeasures: passive forensics pays attention to the task of Deepfake detection while the pristine face is non-protected; proactive defense concentrates on the task of Deepfake destruction; and our proactive forensics focuses on both the tasks of source tracing and Deepfake detection.}\label{fig:countermeasures}
	\Description{Illustration of different types of Deepfake countermeasures: passive forensics, proactive defense, and our proactive forensics.}
\end{figure}
\begin{figure*}[t]
	\centering
	\includegraphics[width=\linewidth]{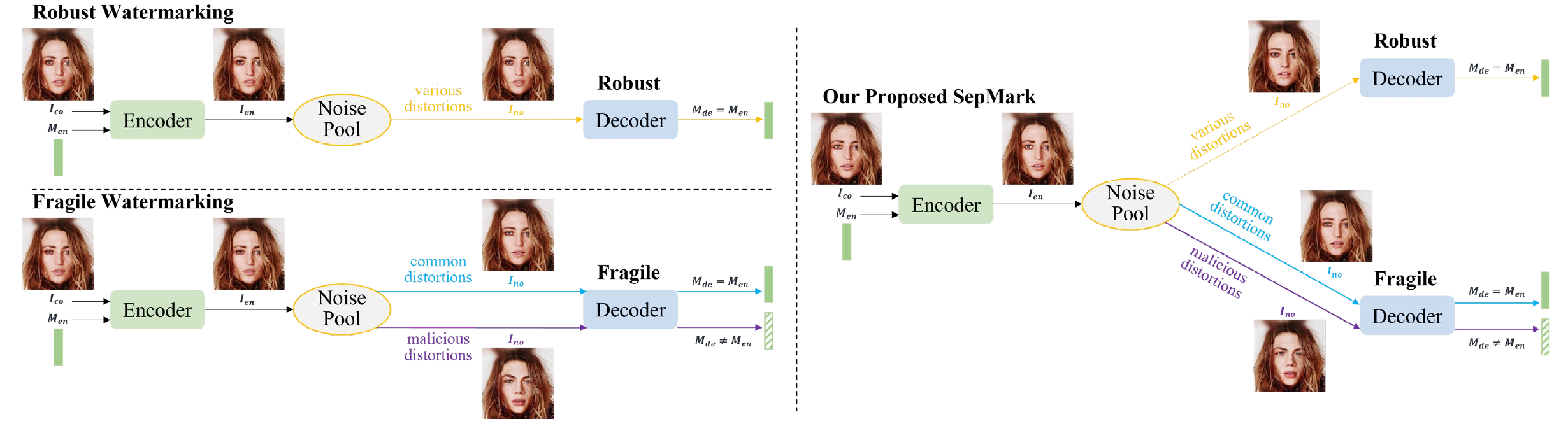}
	\caption{The pipelines of robust watermarking (left top), fragile watermarking (left bottom), and our proposed SepMark (right). For comparison and clarity, the additional adversary discriminator is omitted here. In the prior encoder-decoder structure, an encoder and a decoder are jointly trained by sampling various distortions from the noise pool, where the robust decoder can resist against all of the distortions and extract the watermark in a high level of robustness; alternatively, the fragile one is selectively sensitive to malicious distortions and the watermark can only be extracted in a lower level of robustness. SepMark innovatively integrates a single encoder with two separable decoders, enabling the embedded watermark to be extracted separately at different levels of robustness.}\label{fig:watermarks}
	\Description{Pipelines of robust watermarking, fragile watermarking, and our SepMark.}
\end{figure*}

Until now, AI-generated content has been fully fueled in the deep learning community, such as Generative Pre-trained Transformer \cite{radford2018improving, radford2021learning} and Stable Diffusion \cite{rombach2022high}. In this era, everyone can be a master manipulator who creates authentic-seeming content with just one click \cite{aneja2022tafim}. However, the breakthrough progress in multimedia generation has once again brought Deepfake, which mainly refers to the technology of faking face using deep learning \cite{sun2022fenerf, xu2022mobilefaceswap, xu2022styleswap}, to the forefront. Generally speaking, the development of the forgery generation is usually ahead of its forensic counterpart in time \cite{juefei2022countering, dolhansky2020deepfake}. Despite its significance in face de-identification \cite{cao2021personalized, li2023riddle} and privacy protection \cite{mirjalili2020privacynet, yuan2022generating}, some individuals intentionally produce and spread Deepfakes with high visual deception to distort the truths and manipulate public opinions. In the context of ``seeing is not always believing'', uncontrollable Deepfakes will eventually lead to unprecedented trust crisis and moral panic. Therefore, the forensics of Deepfakes is becoming increasingly important and urgent.

Passive forensics has been introduced since the emergence of Deepfakes \cite{zhou2017two}. Hand-craft features such as eye blink \cite{li2018ictu}, head pose \cite{yang2019exposing}, color inconsistency \cite{li2020identification}, etc. work well for detecting them, at an early stage. However, more clever creators using adversarial training can readily fix these shallow manipulation artifacts \cite{ding2021anti, huang2020fakepolisher}. Vanilla deep-learned features were demonstrated to be effective \cite{rossler2019faceforensics++}, while suffering from generalization when encountering unseen forgery methods. Therefore, some data-driven models aim to learn common artifacts, such as blending boundary \cite{li2020face}, spectral anomaly \cite{durall2020watch}, temporal inconsistency \cite{hu2022finfer}, etc., while avoiding overfitting specific forgery method. Nevertheless, their performances when the inquiry face undergoes several post-processing operations or adversarial perturbations are unclear \cite{carlini2020evading, liao2023famm, hu2021detecting, hussain2021adversarial, neekhara2021adversarial}, rendering that the in-the-wild forensics remains a challenge due to limited robustness \cite{huang2021dodging, hou2023evading, zi2020wilddeepfake}. Despite a handful of proactive defense methods using adversarial attacks have been dedicated to disrupting or nullifying the creation of Deepfakes \cite{aneja2022tafim, he2022defeating, huang2021initiative, huang2022cmua, ruiz2020disrupting, wang2022anti, yeh2020disrupting, yeh2021attack}, they may contribute to the detection task \cite{wang2022deepfake, chen2021magdr}, but not to the source tracing task \cite{lin2022source, pei2021vision, wang2021faketagger} which is another focus of our proactive forensics. We highlight the differences between these countermeasures in Fig.~\ref{fig:countermeasures}, and the application of proactive forensics will be detailed later in Section~\ref{sec:application}.

To our knowledge, invisible watermarking is the art of imperceptibly concealing additional messages within valuable media without influencing their nature-looking \cite{cox2007digital}. According to whether the original image is available during watermark extraction, invisible watermarking can be further categorized as blind and non-blind. Since the extraction needs only the marked image, blind watermarking has increasingly broader applications in modern multimedia ecology \cite{chang2022blind, fu2022chartstamp, yu2021artificial}, including copyright protection, source tracing, content authentication, etc. Therefore, we leverage blind watermarking to combat Deepfakes in this paper. Thanks to the efforts made by deep learning in blind watermarking \cite{zhu2018hidden}, encoder-decoder-based deep watermarking has attracted more interest owing to its effectiveness and flexibility \cite{jia2021mbrs, neekhara2022facesigns}, compared to traditional watermarking \cite{ni2006reversible, wu2022sign}. Still, in accordance with Fig.~\ref{fig:watermarks}, let us conduct an in-depth analysis of why current deep watermarking with single-function cannot meet the proactive forensics well. 

Robust watermarking places a high priority on robustness, emphasizing that the embedded watermark can still be extracted after severe distortions \cite{zhu2018hidden}. This is crucial to protect the copyright and trace the source, even though the marked image has major portions occluded or has been captured by a screenshot \cite{tancik2020stegastamp, jia2022learning}. However, the overly high level of robustness may also allow the watermark to survive both before and after Deepfake, making it difficult to distinguish between genuine and forged images. To make matters worse, the original source identification is preserved even after the image has been forged. Therefore, relying solely on robust watermarks cannot detect Deepfakes while tracing their trusted sources. 

Fragile watermarking focuses on full or selective fragility, which is opposite to robust watermarking, and emphasizes that the embedded watermark will change accordingly with the marked image under any or specific distortions \cite{asnani2022proactive}. Indeed, it mainly indicates whether the content has been altered by malicious distortions, but cannot provide information about the source of original trustworthy image due to the disappeared watermark \cite{neekhara2022facesigns}. More importantly, it is hard to tell apart the forged image from pristine images without prior knowledge, since all of them have no correct watermark \cite{sun2022faketracer}. Therefore, solely relying on fragile watermarks cannot always trace the trusted source or detect Deepfakes.

To accomplish both the tasks of source tracing and Deepfake detection, the most intuitive manner is to embed one robust watermark followed by another fragile watermark \cite{shen2012single}, while the two watermarks may significantly affect each other and concurrently degrade the visual quality. Another intuition is to embed these independent watermarks into separate regions of interest, foreground and background of the face image, for example, while we argue that the division of the regions may require the participation of extra sophisticated face parsing models. More importantly, the watermarks are more likely to be perceived and attacked by virtue of the knowledge of the interest regions. To remedy these limitations of multi-embedding, a learning-based, end-to-end SepMark is proposed as a whole, corresponding to the new paradigm where one elegant embedding has two distinct extraction manners. To conclude, we have the following remarkable contributions:

1) We propose for the first time the deep separable watermarking, SepMark, which brings a new paradigm to current deep watermarking, where a single encoder embeds one watermark elegantly, while two decoders can extract the watermark separately at different levels of robustness. On top of that, our proposed SepMark also naturally provides a novel proactive forensics solution for unified source tracing and Deepfake detection.

2) This is the very first end-to-end learning architecture for deep separable watermarking, which consists of a single encoder, a discriminator, and two separable decoders trained by sampling different types of distortions from a random forward noise pool (RFNP). The robust decoder resists against all the common and malicious distortions while the semi-robust one is selectively sensitive to malicious distortions.

3) Extensive experiments conducted on face images demonstrate the high level of robustness of robust decoder under numerous distortions, and the selective fragility of semi-robust decoder under malicious Deepfake distortions. Therefore, the separable decoders can reliably trace the trusted source of the marked face and detect whether it has been altered since being marked.

\section{Related Work}

\subsection{Deepfake Passive Forensics}
Passive forensics, especially self-supervised Deepfake detection, has recently gained much attention \cite{zhao2022self}. The fundamental principle among them is the synthesis of self-craft Deepfakes driven by positive examples. In general, their detection performances depend on the prior knowledge introduced during the production of synthetic data. FWA \cite{li2018exposing} targets the artifacts of affine face warping due to inconsistent resolutions between inner and outer faces. Face X-ray \cite{li2020face} focuses on the artifacts of blending boundary that reveal whether the inquiry image can be decomposed into the blending of two faces from different sources. Similarly, I2G \cite{zhao2021learning} aims to generate the blended faces with inconsistent source features through a blending pipeline. SBIs \cite{shiohara2022detecting} applies two random transformations to the single image, one as the source and the other as the target of the blending process, to reproduce more challenging negative samples. Moreover, ICT \cite{dong2022protecting} considers the identity consistency in the inquiry image, such that the inner face and outer face of different identities are swapped to form synthetic data. Following these additional cues \cite{chen2022self, hong2021self, zhou2022detection}, more general modeling of Deepfake distortions can take a step forward towards generalizing to unseen forgeries.

\subsection{Deepfake Proactive Defense}
Similar to proactive forensics, proactive defense requires the addition of imperceptible signals to the pristine image in advance. Deepfake destruction can be achieved either by disrupting or nullifying the Deepfake model through the injection of adversarial perturbations. The former can be regarded as a non-targeted attack, where the attacked image after Deepfake should be visually abnormally and as unnatural as possible compared to the original forged one \cite{huang2021initiative, huang2022cmua, ruiz2020disrupting, wang2022anti, yeh2020disrupting}. Conversely, the latter is a targeted attack, where the attacked image after Deepfake should be as similar as possible to the original pristine one \cite{he2022defeating, yeh2021attack}. Inspired by universal adversarial perturbations \cite{moosavi2017universal}, CMUA \cite{huang2022cmua} generates image-universal cross-model perturbations to attack multiple Deepfake models rather than the specific one. Anti-Forgery \cite{wang2022anti} further enhances the robustness of perceptual-aware perturbations against various distortions. However, the visually perceptible artifacts might still escape the detection since passive detectors do not share a similar perspective as human eyes \cite{wang2022deepfake}. Moreover, counterparts like MagDR \cite{chen2021magdr} can detect and remove the adversarial perturbations; even the Deepfakes can be further reconstructed normally.

\subsection{Deep Watermarking}
Motivated by the success of deep learning in computer vision tasks, DNN-based models have been investigated in blind watermarking for stronger robustness. HiDDeN \cite{zhu2018hidden} is the first end-to-end learning architecture for robust watermarking, which consists of an encoder, a discriminator, a noise layer and a decoder. The key to guaranteeing robustness is the data augmentation of the marked image with the differential noise layer. To resist against non-differential JPEG distortion, MBRS \cite{jia2021mbrs} randomly switches real JPEG and simulated differential JPEG for each mini-batch training. CIN \cite{ma2022towards} incorporates an invertible neural network to learn a joint representation between watermark embedding and extraction, where a non-invertible attention-based module is additionally devised against non-additive quantization noise. In another distortion-agnostic line \cite{luo2020distortion}, TSDL \cite{liu2019novel} proposes a two-stage learning architecture, where both the encoder and decoder are first trained in a noise-free manner, following that only the decoder is fine-tuned to adapt to black-box distortions. ASL \cite{zhang2021towards} models the forward noise as pseudo-noise that interacts with the marked image, allowing the gradient to be propagated to the encoder without passing through the noise layer during backward propagation. FIN \cite{fang2022flow} handles the black-box distortion through an invertible noise layer similar to invertible denoising \cite{liu2021invertible}; it is designed to simulate the distortion, serves as a noise layer during training, and is also applied as a preprocessing step before extraction to eliminate the distortion. Moreover, LFM \cite{wengrowski2019light}, StegaStamp \cite{tancik2020stegastamp}, RIHOOP \cite{jia2020rihoop}, and PIMoG \cite{fang2022pimog} learn to resist against physical distortions like printer-scanner, printer-camera, and screen-camera that are cross-media. To our knowledge, although deep robust watermarking has been developed a lot, deep fragile watermarking remains largely unexplored.

\begin{figure}[t]
	\centering
	\includegraphics[width=\linewidth]{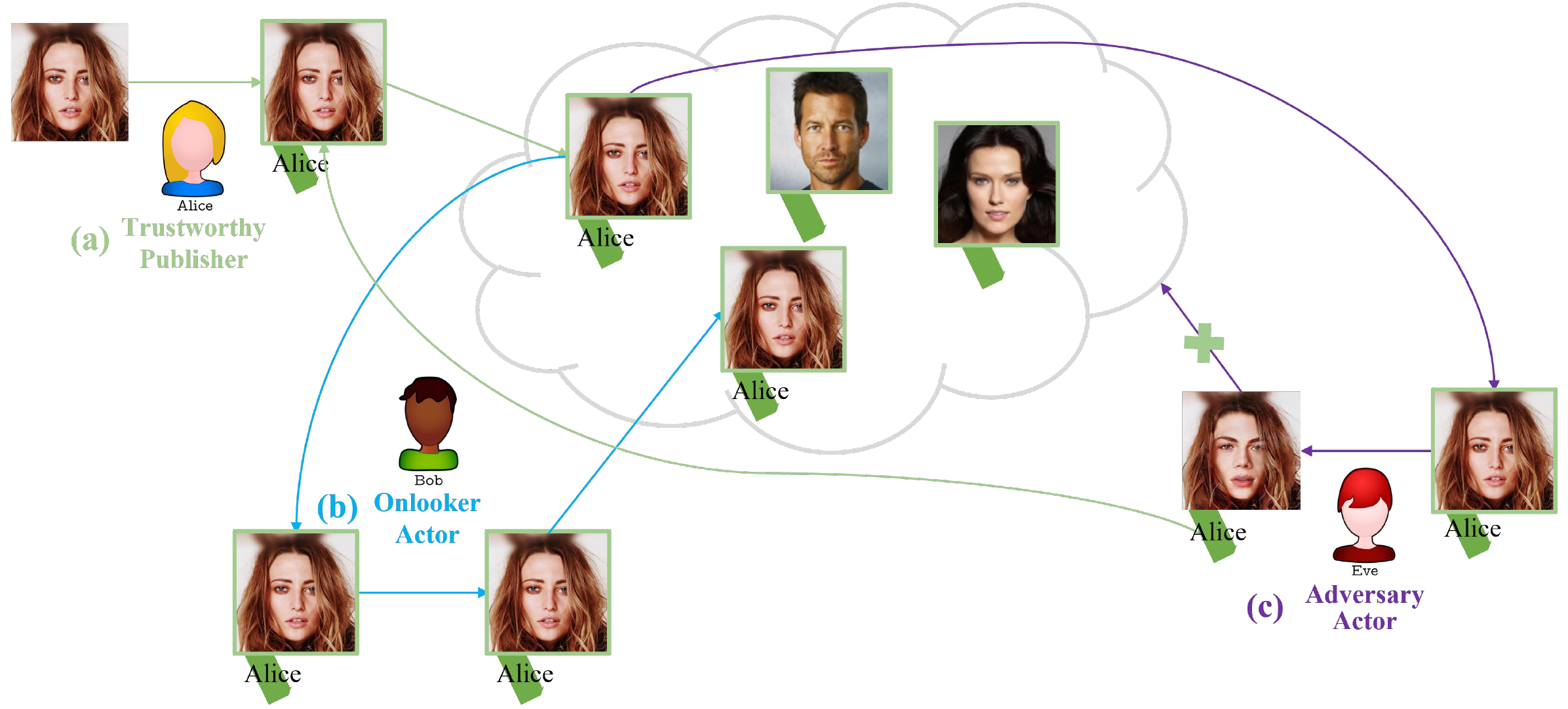}
	\caption{Illustrative example of proactive forensics. (a) Trustworthy publisher Alice embeds her identification into the published face. (b) Onlooker actor Bob reposts this face without affecting the content semantics. (c) Adversary actor Eve manipulates this face to distort the truth.}\label{fig:example}
	\Description{Illustrative example.}
\end{figure}
\begin{figure*}[t]
	\centering
	\includegraphics[width=\linewidth]{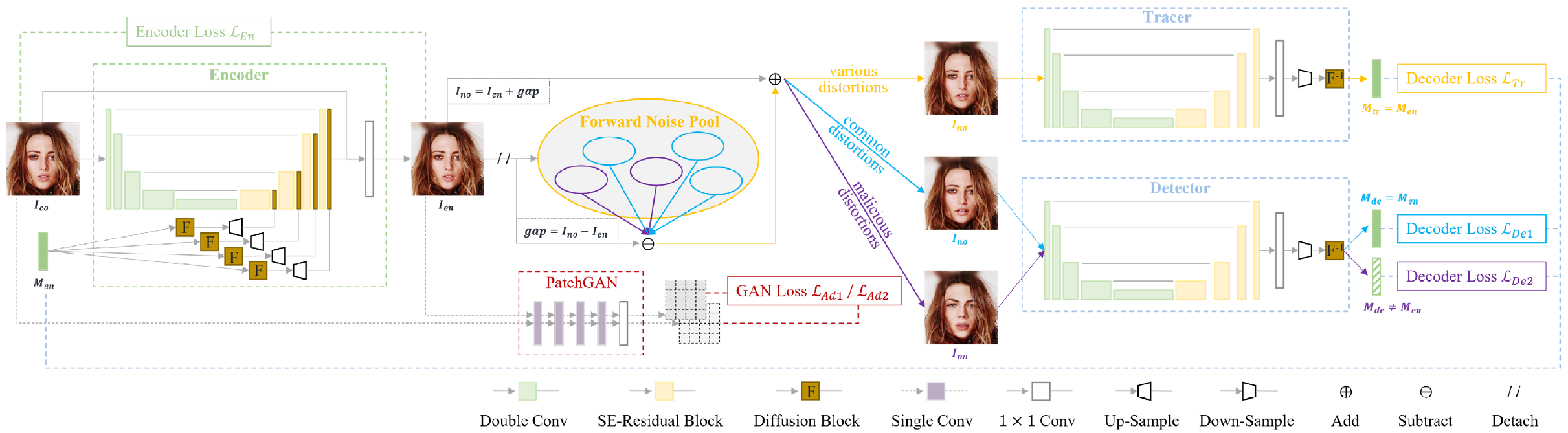}
	\caption{Overall architecture of SepMark. The encoder $En$ receives the cover image $I_{co}$ and encoded message $M_{en}$ as input and produces the encoded image $I_{en}$. The random forward noise pool $RFNP$ samples different types of pseudo-noise $gap$ in a detached forward propagation and interacts with $I_{en}$ in both standard forward and backward propagation, resulting in the common or maliciously distorted image $I_{no}$. The tracer $Tr$ extracts the message $M_{tr}$ which is identical to $M_{en}$ from the arbitrary distorted image. The detector $De$ extracts the correct message from the common distorted image but the wrong message from the maliciously distorted image. Additionally, a PatchGAN $Ad$ is used to distinguish each pristine and encoded image patch.}\label{fig:architecture}
	\Description{Overall architecture.}
\end{figure*}

\section{Proactive Forensics: SepMark}

\subsection{Problem Statement}\label{sec:application}
Instead of detecting which are forged afterwards, SepMark is dedicated to ensuring beforehand which are genuine and tracing their trusted sources. This proactive forensics should work on the pristine face in advance, where ``pristine'' means the face has not been processed by SepMark in this paper. In the real world, a trustworthy content publisher uploads his/her pristine face to social networks and makes it public to anyone. Unfortunately, some adversary individuals may download the face and manipulate it to spread misinformation, which causes severe impacts on social trust and public interest. Moreover, relying on ex-post forensics, the spread of Deepfakes cannot be immediately blocked before it is confirmed.

Here is a more illustrative example regarding the suggested proactive forensics shown in Fig.~\ref{fig:example}. To provide preventive measures and more reliable evidence for the pristine face, SepMark embeds one watermark into the face in advance (Fig.~\ref{fig:example}-a), such as the publisher identification and time stamp, which can identify the trusted source of the marked face and at the same time indicate it has been processed by SepMark. Subsequently, the embedded watermark can be extracted separately at different levels of robustness. Concretely, both the robust and semi-robust watermarks can be extracted correctly under common distortions that have little effect on image semantics (Fig.~\ref{fig:example}-b), such as JPEG compression, resizing, contrast adjustments, etc. In contrast, under malicious distortions such as Deepfake, which is mainly addressed here (Fig.~\ref{fig:example}-c), the robust watermark can still be extracted correctly, while the semi-robust one is extracted incorrectly, implying that the face image has been altered since it was marked. In this manner, any individuals or social networks can verify whether the original source of the face comes from a trustworthy publisher. Furthermore, even without the knowledge of original watermark, it is possible to compare the two extracted watermarks to determine whether the marked face has been manipulated after publication, and block the further spreading of the suspicious image with mismatched watermarks. In summary, SepMark accomplishes both provenance tracking and blind detection of Deepfake without knowing the original image.

Following Fig.~\ref{fig:architecture}, let us formulate SepMark in detail. Given an arbitrary pristine face which also refers to as the cover image $I_{co}$, the encoder $En$ embeds encoded message $M_{en}$ into the image $I_{co}$, thus obtaining an encoded image $I_{en}$. After that, the separable Tracer $Tr$ and Detector $De$ can extract the messages $M_{tr}$ and $M_{de}$ from the common or maliciously distorted image $I_{no}$, respectively. Consequently, depending on the separable decoders $Tr$ and $De$, we can divide the proactive forensics into the following cases.

Case 1: If the robust decoder $Tr$ is available, the extracted $M_{tr}$ is expected to always be identical to the embedded $M_{en}$, whether the marked image is noised by common or malicious distortions, to reliably trace the trusted source.

Case 2: If the semi-robust decoder $De$ is available, it can extract the correct message from the common distorted image but the wrong message from the image noised by malicious distortions, to distinguish the type of the distortion.

Case 3: Only when both the decoders $Tr$ and $De$ are available, the tracing of the trusted source and the detection of malicious distortions can be realized simultaneously. More specifically, if $M_{de}\approx M_{en}\text{ or } M_{de}\approx M_{tr}$, SepMark will indicate that the face is genuine and comes from the trusted source $M_{tr}$. Otherwise, if $M_{de}\napprox M_{en}\text{ or } M_{de}\napprox M_{tr}$, the face has been forged since it was marked, and the original trustworthy one can be further acquired from the source $M_{tr}$ if necessary. 

\subsection{Model Architecture}
To achieve end-to-end training of deep separable watermarking, our proposed SepMark consists of a single encoder $En$, a random forward noise pool $RFNP$, two separable decoders $Tr$ and $De$, and an additional adversary discriminator $Ad$, as can be seen in Fig.~\ref{fig:architecture}. In a nutshell, 1) $En$ receives a batch of $I_{co}\in \mathbb{R}^{B\times 3\times H\times W}$ and a batch of $M_{en}\in \{-\alpha, \alpha\}^{B\times L}$ to produce $I_{en}$; 2) $RFNP$ randomly samples different distortions as pseudo-noise and interacts with $I_{en}$, resulting in $I_{no}$; 3) $Tr$ and $De$ separately extract $M_{tr}$ and $M_{de}$ from $I_{no}$ with different levels of robustness; and 4) $Ad$ classifies pristine or encoded for each image patch of $I_{en}$ relative to $I_{co}$.

\subsubsection{Single Encoder}
In the main line, $I_{co}$ is fed into a UNet-like structure. Firstly, in the downsampling stage, ``conv-in-relu'' with stride $2$ halves the spatial dimensions, and ``Double Conv ($2\times$ conv-in-relu)'' doubles the number of feature channels. In the parallel lines, $M_{en}$ is fed into several ``Diffusion Block (FC layer-Conv)'' to inject redundancy, where its length $L$ is expended to $L\times L$. Secondly, in the upsampling stage, nearest-neighbour interpolation doubles the spatial dimensions and ``conv-in-relu'' halves the number of feature channels, followed by the concatenation with both the corresponding feature maps from the downsampling stage and the redundant message after interpolation operation at each resolution. Moreover, ``SE-Residual Block'' inserted with Squeeze-and-Excitation \cite{hu2018squeeze} replaces batch normalization in the original bottleneck residual block with instance normalization, and is applied to the concatenated features to enhance the information interaction between different channels. Finally, we further concatenate the aforementioned feature maps with the cover image via a skip connection, and fed the output into a $1\times 1$ convolution to produce the encoded image $I_{en}$.

\subsubsection{Random Forward Noise Pool}
Considering that numerous distortions and a specific noise layer cannot adapt to all of them, it is proposed to introduce various distortions to our noise pool for the combined training. Moreover, given that the simulated distortion struggles to approach the real complex distortion, we directly get arbitrary real distortion $gap=I_{no}-I_{en}$ in a detached forward propagation. In this way, both the standard forward and backward propagation can be realized through differentiable $I_{no}=I_{en}+gap$. It is noteworthy that the difference between $RFNP$ and ASL \cite{zhang2021towards} lies in the random sampling of different distortions in each mini-batch while not that of only one distortion. Empirically, this change benefits our stable training and explains the instance normalization in our encoder design. To be specific, random sampling from common and malicious distortions, common distortions only, and malicious distortions only results in three batches of $I_{no}$.

\subsubsection{Separable Decoders}
We simply borrow the UNet-like structure from our encoder for the two decoders. Note that after the concatenations of corresponding feature maps at each resolution, we fed the final output into a $1\times 1$ convolution to produce a single-channel residual image. After that, the residual image is downsampled to $L\times L$, and an ``Inverse Diffusion Block (FC layer)'' is used to extract the decoded message of length $L$. More importantly, due to different inputs and optimization objectives during training, the decoders have the same structural design but do not share model parameters; see our carefully designed loss functions later. To elaborate further, the robust $Tr$ receives the image $I_{no}$ noised by various distortions, while the semi-robust $De$ receives the common and maliciously distorted ones, respectively, in each mini-batch training. Moreover, $Tr$ is optimized by the objective $M_{tr}=M_{en}$ under arbitrary distortions. In comparison, $De$ is optimized by the objective $M_{de}=M_{en}$ under common distortions, and the objective $M_{de}\neq M_{en}$ under malicious distortions.

\subsubsection{Adversary Discriminator}
To supervise the visual quality of encoded image $I_{en}$, we also resort to adversarial training, where an additional discriminator $Ad$ has trained alternately with the main encoder-decoder. In particular, vanilla GAN is replaced with PatchGAN \cite{isola2017image, zhu2017unpaired} to classify each image patch as pristine or encoded. Specifically, the discriminator is a fully-convolutional structure consisting of several ``conv-in-leakyrelu'' blocks, and finally, a $1\times 1$ convolution outputs the ultimate prediction map. However, the GAN training process was found usually unstable and even non-convergence when using the standard GAN loss \cite{goodfellow2020generative}. This behavior also appears in our training process, and we conjecture that it might also be related to the task of message extraction, in addition to adversarial generation and discrimination. To improve the stability of the training, we use the GAN loss of RaLSGAN \cite{jolicoeur2018relativistic} instead, and we find that this adjustment makes our training more stable. This might be mainly attributed to the transformation from absolute true or false into relative true or false.

\subsection{Loss Functions}

\begin{figure*}[t]
	\centering
	\includegraphics[width=0.82\linewidth]{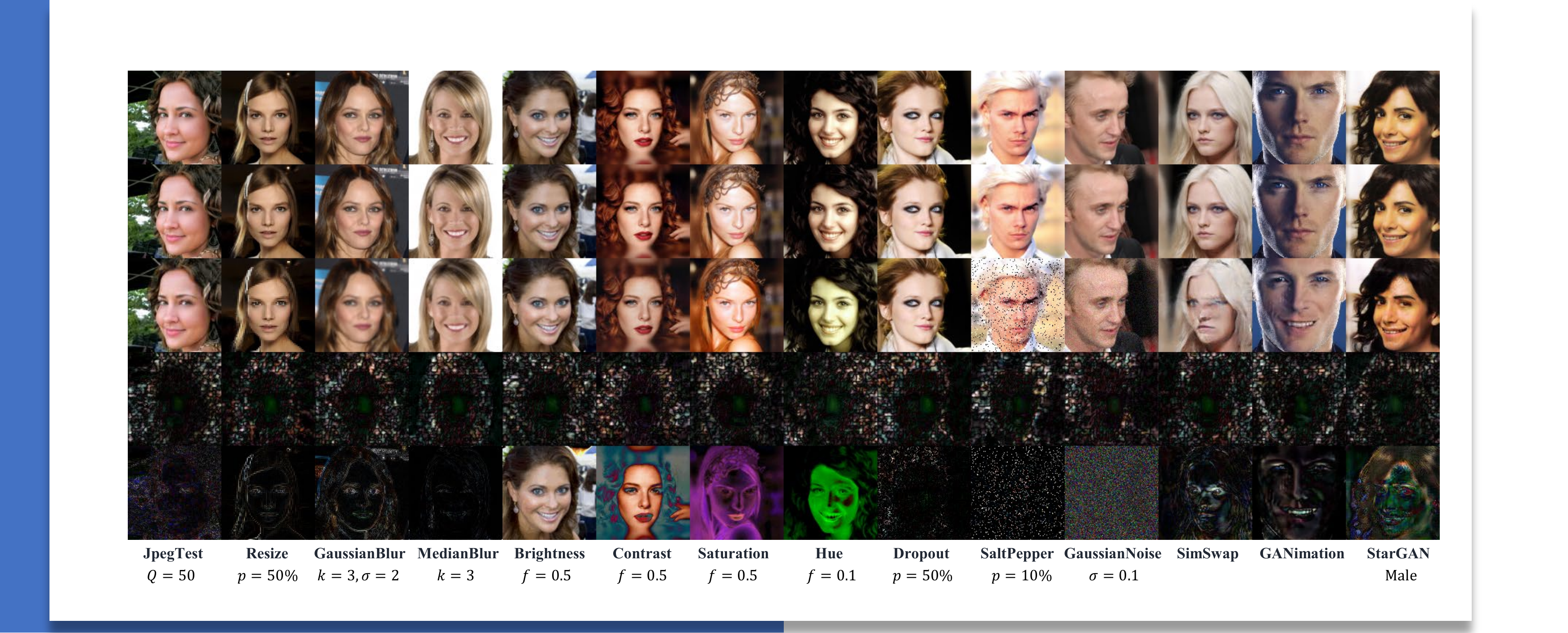}
	\caption{Subjective visual quality under typical distortions. From top to bottom are the cover image $I_{co}$, the encoded image $I_{en}$, the noised image $I_{no}$, the residual signal of $\mathcal{N}(|I_{co}-I_{en}|)$, and $\mathcal{N}(|I_{no}-I_{en}|)$, where $\mathcal{N}(I)=(I-min(I))/(max(I)-min(I))$. Each column corresponds to one distortion. Image size: $128\times 128$. For more malicious distortions, please see Appendix~\ref{sec:additional_visual_examples}.}\label{fig:subjective}
	\Description{Subjective visual quality.}
\end{figure*}

We found simple $L_{2}$ constraint is sufficient for the end-to-end training of SepMark. Note that $\theta$ indicates trainable parameters, and $\theta_{co}$ and $\theta_{ma}$ only denote common and malicious distortions sampled by $RFNP$, respectively. We have the following loss functions.

For the encoder, its encoded image $I_{en}$ should be visually similar to the input image $I_{co}$:
\begin{equation}
	\mathcal{L}_{En}=L_{2}(I_{co},En(\theta,I_{co},M_{en}))=L_{2}(I_{co},I_{en})
\end{equation}

For the decoders, the robust Tracer $Tr$ should be able to extract $M_{tr}$, which is identical to $M_{en}$, from the image $I_{no}$ noised by arbitrary common and malicious distortions:
\begin{equation}
	\mathcal{L}_{Tr}=L_{2}(M_{en},Tr(\theta,RFNP(\theta_{co} + \theta_{ma}, I_{en})))=L_{2}(M_{en},M_{tr})
\end{equation}
where
\begin{equation}
	I_{en}=En(\theta,I_{co},M_{en})
\end{equation}
The semi-robust Detector $De$ is expected to be fault-tolerant towards the common distorted image $I_{no}$:
\begin{equation}
	\mathcal{L}_{De1}=L_{2}(M_{en},De(\theta,RFNP(\theta_{co}, I_{en})))=L_{2}(M_{en},M_{de})
\end{equation}
In contrast, $De$ should be selectively sensitive to malicious distortions $\theta_{ma}$:
\begin{equation}
	\mathcal{L}_{De2}=L_{2}(0,De(\theta,RFNP(\theta_{ma}, I_{en})))=L_{2}(0,M_{de})
\end{equation}
where the $L_{2}$ distance between $0$ and $M_{de}$ is optimized. It should be emphasized that the binary message $\{0, 1\}$ had been transferred to the zero-mean $\{-\alpha, \alpha\}$. Therefore, if the extracted message is greater than $0$, it is recovered as binary $1$, otherwise as binary $0$. This loss term makes the extracted $M_{de}$ close to random guesses.

For the discriminator, its parameters are updated by:
\begin{equation}
	\mathcal{L}_{Ad1}=L_{2}(Ad(\theta,I_{co})-\overline{Ad}(\theta,I_{en}),1)+L_{2}(Ad(\theta,I_{en})-\overline{Ad}(\theta,I_{co}),-1)
\end{equation}
where 
\begin{equation}
	\overline{Ad}(I)=\frac{1}{B}\sum_{i=1}^{B}Ad(I^{i\times 3\times H\times W}) %mean(Ad(I))
\end{equation}
\begin{equation}
	Ad(\theta,I_{en})=Ad(\theta,En(I_{co},M_{en}))
\end{equation}
The relativistic average discriminator here estimates the probability that the cover image $I_{co}$ is more pristine than the encoded image $I_{en}$ on average; please refer to the literature \cite{jolicoeur2018relativistic} for more details. Note that the discriminator also participates in updating the encoder $En$:
\begin{equation}
	\mathcal{L}_{Ad2}=L_{2}(Ad(I_{co})-\overline{Ad}(I_{en}),-1)+L_{2}(Ad(I_{en})-\overline{Ad}(I_{co}),1)
\end{equation}
where
\begin{equation}
	Ad(I_{en})=Ad(En(\theta,I_{co},M_{en}))
\end{equation}

To summarize, the total loss for the main encoder-decoder can be formulated by:
\begin{equation}
	\mathcal{L}_{Total}=\lambda_{1}\mathcal{L}_{Ad2}+\lambda_{2}\mathcal{L}_{En}+\lambda_{3}\mathcal{L}_{Tr}+\lambda_{4}\mathcal{L}_{De1}+\lambda_{5}\mathcal{L}_{De2}\label{eq:11}
\end{equation}
where $\lambda_{1}$, $\lambda_{2}$, $\lambda_{3}$, $\lambda_{4}$, $\lambda_{5}$ are the weights for respective loss terms.

\section{Experiments}

\subsection{Implementation Details}
The experiments mainly conduct on CelebA-HQ \cite{karras2017progressive, lee2020maskgan}, where $24183/2993/2824$ face images are used for training, valuation, and testing, referencing the official splits of CelebA \cite{liu2015deep}. Additionally, the testing set of CelebA, which includes $19962$ face images, and the valuation set of COCO \cite{lin2014microsoft}, which contains $5000$ images of common objects, are employed to evaluate the generalizability. Unless stated otherwise, all the images are resized to two resolutions of $128\times 128$ and $256\times 256$, due to the limit of computation resource. For the malicious distortions, SimSwap \cite{chen2020simswap}, GANimation \cite{pumarola2018ganimation}, and StarGAN \cite{choi2018stargan} are selected as they are typical Deepfakes, which belong to representative face swapping, expression reenactment, and attribute editing, respectively. In detail, the target face used for swapping is randomly selected from the valuation set of CelebA, the target expression used for reenactment is randomly selected from the driving images ``eric\_andre'' \footnote{\scriptsize https://github.com/vipermu/ganimation/tree/master/animations/eric\_andre/attribute\_images}, the attribute set we edit is $\{Male,Young,BlackHair,BlondHair,BrownHair\}$, and all of them are implemented based on their released pre-trained model for resulting convincing Deepfakes. Note that the face images are further background-removed by OpenFace \cite{baltrusaitis2018openface} only when testing GANimation, to match the requirements of the input. Without loss of generality, the set of common distortions is $\{Identity,JpegTest,Resize,\\GaussianBlur,MedianBlur,Brightness,Contrast,Saturation,Hue,\\Dropout,SaltPepper,GaussianNoise\}$. The detailed parameters of the above distortions can be seen at the bottom of Fig.~\ref{fig:subjective}. Note that cropping is not included in the set since it is difficult to understand as a common distortion for the complete face.

\begin{table}[t]
	\caption{Objective visual quality of the encoded image $I_{en}$.}
	\label{tab:objective}
	\small
	\renewcommand\arraystretch{0.9}
	\begin{tabular}{lccccc}
		\toprule
		Model&\begin{tabular}[c]{@{}c@{}}Image\\ Size\end{tabular}&\begin{tabular}[c]{@{}c@{}}Message\\ Length\end{tabular}&PSNR $\uparrow$ &SSIM $\uparrow$ &LPIPS $\downarrow$\\
		\midrule
		MBRS \cite{jia2021mbrs} & $128\times 128$ & 30 & 33.0456 & 0.8106 & 0.0141\\
		CIN \cite{ma2022towards} & $128\times 128$ & 30 & 42.4135 & 0.9628 & 0.0006\\
		PIMoG \cite{fang2022pimog} & $128\times 128$ & 30 & 37.7271 & 0.9470 & 0.0086\\
		SepMark & $128\times 128$ & 30 & 38.5112 & 0.9588 & 0.0028\\
		\hline
		FaceSigns \cite{neekhara2022facesigns} & $256\times 256$ & 128 & 32.3319 & 0.9211 & 0.0260\\
		SepMark & $256\times 256$ & 128 & 38.5646 & 0.9328 & 0.0080\\
		\bottomrule
	\end{tabular}
\end{table}

\begin{table*}[t]
	\caption{Robustness test for the common distorted image $I_{no}$, where ``Tracer'' means the BER between $M_{tr}$ and $M_{en}$, ``Detector'' means the BER between $M_{de}$ and $M_{en}$, and ``Both'' means the BER between $M_{tr}$ and $M_{de}$.}
	\label{tab:common_distortion}
	\small
	\renewcommand\arraystretch{0.9}
	\begin{tabular}{lcccccccccccc}
		& \multicolumn{6}{c|}{$128\times 128$} & \multicolumn{4}{c}{$256\times 256$}\\
		\toprule
		\multirow{2}{*}{Distortion}&\multirow{2}{*}{MBRS \cite{jia2021mbrs}}&\multirow{2}{*}{CIN \cite{ma2022towards}}&\multirow{2}{*}{PIMoG \cite{fang2022pimog}}&\multicolumn{3}{c|}{SepMark}&\multirow{2}{*}{FaceSigns \cite{neekhara2022facesigns}}&\multicolumn{3}{c}{SepMark}\\ \cline{5-7} \cline{9-11}
		&&&&{Tracer}&Detector&\multicolumn{1}{c|}{Both}&&{Tracer}&{Detector}&Both\\
		\midrule
		Identity &0.0000\%&0.0000\%&0.0366\%&0.0000\%&0.0000\%&\multicolumn{1}{c|}{0.0000\%}&0.0136\%&0.0000\%&0.0000\%&0.0000\%\\
		JpegTest &0.2597\%&2.7514\%&19.5562\%&0.2136\%&0.2172\%&\multicolumn{1}{c|}{0.2656\%}&0.8258\%&0.0075\%&0.0069\%&0.0133\%\\
		Resize &0.0000\%&0.0000\%&0.0767\%&0.0059\%&0.0212\%&\multicolumn{1}{c|}{0.0201\%}&1.0726\%&0.0000\%&0.0000\%&0.0000\%\\
		GaussianBlur &0.0000\%&22.7786\%&0.1169\%&0.0024\%&0.0035\%&\multicolumn{1}{c|}{0.0012\%}&0.1671\%&0.0000\%&0.0274\%&0.0274\%\\
		MedianBlur &0.0000\%&0.0307\%&0.0992\%&0.0012\%&0.0012\%&\multicolumn{1}{c|}{0.0000\%}&0.0977\%&0.0000\%&0.0000\%&0.0000\%\\
		Brightness &0.0000\%&0.0000\%&1.3443\%&0.0059\%&0.0106\%&\multicolumn{1}{c|}{0.0142\%}&10.8196\%&0.0017\%&0.0030\%&0.0047\%\\
		Contrast &0.0000\%&0.0000\%&0.8121\%&0.0012\%&0.0024\%&\multicolumn{1}{c|}{0.0012\%}&0.0334\%&0.0000\%&0.0000\%&0.0000\%\\
		Saturation &0.0000\%&0.0000\%&0.0803\%&0.0000\%&0.0000\%&\multicolumn{1}{c|}{0.0000\%}&0.7113\%&0.0000\%&0.0000\%&0.0000\%\\
		Hue &0.0000\%&0.0000\%&0.1523\%&0.0000\%&0.0012\%&\multicolumn{1}{c|}{0.0012\%}&8.3780\%&0.0000\%&0.0000\%&0.0000\%\\
		Dropout &0.0000\%&0.0000\%&0.4828\%&0.0000\%&0.0000\%&\multicolumn{1}{c|}{0.0000\%}&17.5615\%&0.0058\%&0.0000\%&0.0058\%\\
		SaltPepper &0.0000\%&0.0378\%&2.3667\%&0.0413\%&0.0106\%&\multicolumn{1}{c|}{0.0472\%}&12.3238\%&0.0008\%&0.0003\%&0.0011\%\\
		GaussianNoise &0.0000\%&0.0000\%&12.7396\%&0.7460\%&0.8735\%&\multicolumn{1}{c|}{0.8994\%}&7.0697\%&0.0578\%&0.0622\%&0.0930\%\\
		\hline
		Average &0.0216\%&2.1332\%&3.1553\%&0.0848\%&0.0951\%&\multicolumn{1}{c|}{0.1042\%}&4.9228\%&0.0061\%&0.0083\%&0.0121\%\\
		\bottomrule
	\end{tabular}
\end{table*}

\begin{table*}[t]
	\caption{Robustness test for the maliciously distorted image $I_{no}$.}
	\label{tab:malicious_distortion}
	\small
	\renewcommand\arraystretch{0.9}
	\begin{tabular}{lcccccccccccc}
		& \multicolumn{6}{c|}{$128\times 128$} & \multicolumn{4}{c}{$256\times 256$}\\
		\toprule
		\multirow{2}{*}{Distortion}&\multirow{2}{*}{MBRS \cite{jia2021mbrs}}&\multirow{2}{*}{CIN \cite{ma2022towards}}&\multirow{2}{*}{PIMoG \cite{fang2022pimog}}&\multicolumn{3}{c|}{SepMark}&\multirow{2}{*}{FaceSigns \cite{neekhara2022facesigns}}&\multicolumn{3}{c}{\hspace{-1.5mm}SepMark}\\  \cline{5-7} \cline{9-11}
		&&&&{Tracer}&Detector&\multicolumn{1}{c|}{Both}&&{Tracer}&{Detector}&Both\\
		\midrule
		SimSwap &19.3744\%&48.5068\%&8.6745\%&13.8255\%&50.8829\%&\multicolumn{1}{c|}{50.8558\%}&49.9463\%&7.9068\%&45.9117\%&45.6713\%\\
		GANimation &0.0000\%&0.0000\%&0.4802\%&0.0000\%&36.7938\%&\multicolumn{1}{c|}{36.7938\%}&45.4172\%&0.0020\%&43.8524\%&43.8527\%\\
		StarGAN (Male) &18.3133\%&67.2568\%&9.2044\%&0.1157\%&52.6003\%&\multicolumn{1}{c|}{52.6027\%}&50.2617\%&0.0033\%&45.4624\%&45.4641\%\\
		StarGAN (Young) &17.0562\%&69.0805\%&8.7465\%&0.1074\%&52.3678\%&\multicolumn{1}{c|}{52.3737\%}&50.3649\%&0.0030\%&45.5319\%&45.5333\%\\
		StarGAN (BlackHair) &19.2233\%&58.7913\%&10.5312\%&0.1416\%&49.2434\%&\multicolumn{1}{c|}{49.2599\%}&50.2576\%&0.0019\%&45.1299\%&45.1296\%\\
		StarGAN (BlondHair) &18.4478\%&72.9733\%&10.3506\%&0.1712\%&48.6084\%&\multicolumn{1}{c|}{48.5930\%}&50.9829\%&0.0022\%&44.6953\%&44.6953\%\\
		StarGAN (BrownHair) &17.6381\%&79.5857\%&9.0675\%&0.0980\%&52.3820\%&\multicolumn{1}{c|}{52.3855\%}&50.7884\%&0.0017\%&45.9950\%&45.9939\%\\
		\hline
		Average &15.7219\%&56.5992\%&8.1507\%&2.0656\%&48.9827\%&\multicolumn{1}{c|}{48.9806\%}&49.7170\%&1.1316\%&45.2255\%&45.1915\%\\
		\bottomrule
	\end{tabular}
\end{table*}

Our SepMark is implemented by PyTorch \cite{paszke2019pytorch} and executed on NVIDIA RTX $4090$. Since we address numerous distortions rather than a specific one, SepMark uses the combined training fashion where the noise pool consists of all the common distortions and the malicious distortions, namely SimSwap, GANimation, and StarGAN (Male). The whole training lasted for $100$ epochs with a batch size of $16$, and as a rule of thumb, we adjusted the Adam optimizer \cite{kingma2014adam} where $lr=0.0002$, $\beta_{1}=0.5$ for stable adversarial training. Besides, $\lambda_{1}$, $\lambda_{2}$, $\lambda_{3}$, $\lambda_{4}$, $\lambda_{5}$ in Eq.~(\ref{eq:11}) are set to $0.1,1,10,10,10$ in our training, respectively. Intuitively, the message range $\alpha$ in our SepMark is set to $0.1$, and we show that $\alpha$ is essential for the trade-off between visual quality and robustness in the ablation study. 

\textbf{Baselines.} As far as we know, this is the first work on deep separable watermarking, we therefore have to adopt robust watermarking methods such as MBRS \cite{jia2021mbrs}, CIN \cite{ma2022towards}, PIMoG \cite{fang2022pimog}, and fragile watermarking method FaceSigns \cite{neekhara2022facesigns} as our baselines. Two independent models, one trained on images of size $128\times 128$ and messages of length $30$, the other trained on images of size $256\times 256$ and messages of length $128$, for comparison with the pre-trained models of baselines to demonstrate the effectiveness of SepMark.

\textbf{Evaluation Metrics.} For the evaluation of objective visual quality, we report the average PSNR, SSIM, and LPIPS \cite{zhang2018unreasonable} of the encoded images throughout the testing set. For the robustness test, the average bit error ratio (BER) is utilized as the default evaluation metric. In our SepMark, the BER of semi-robust Detector under malicious distortions should approach $50\%$; in other cases, the BER of both Tracer and Detector should approach $0\%$. Suppose that the embedded message $M_{E}$ and the extracted message $M_{D}$, formally,
\begin{equation}
	BER(M_{E},M_{D})=\frac{1}{B}\times \frac{1}{L}\times \sum_{i=1}^{B} \sum_{j=1}^{L} (\mathcal{B}(M_{E}^{i\times j})-\mathcal{B}(M_{D}^{i\times j}))\times 100\%
\end{equation}
where
\begin{equation}
	\mathcal{B}(M)=\left\{
	\begin{array}{ll}
		1 & M>0,\\
		0 & M\leq 0.
	\end{array}
	\right.
\end{equation}
Note that we have strictly adjusted $\mathcal{B}(\cdot)$ for the baselines to the same as their original setting, avoiding affecting their performances.

\begin{table}[t]
	\caption{Ablation experiments. Range of message $M_{en}$ where $\alpha \in\{0.15,0.05\}$. Image size: $128\times 128$. Gain: ``$\downarrow$'' means better.}
	\label{tab:message_range}
	\small
	\renewcommand\arraystretch{0.9}
	\setlength{\tabcolsep}{4pt}
	\begin{tabular}{lcccc}
		& \multicolumn{2}{c|}{$\alpha=0.15$} & \multicolumn{2}{c}{$\alpha=0.05$}\\
		\toprule
		\multirow{2}{*}{Distortion}&\multicolumn{2}{c|}{PSNR $=38.1806$}& \multicolumn{2}{c}{PSNR $=40.8058$}\\
		{}&Tracer&\multicolumn{1}{c|}{Detector}& Tracer & Detector\\
		\midrule
		Identity&0.0000\%&\multicolumn{1}{c|}{0.0000\%}&0.0000\%&0.0000\%\\
		JpegTest&0.1062\%&\multicolumn{1}{c|}{0.1204\%}&0.6941\%&0.7495\%\\
		Resize&0.0012\%&\multicolumn{1}{c|}{0.0000\%}&0.0614\%&0.0696\%\\
		GaussianBlur&0.0000\%&\multicolumn{1}{c|}{0.0000\%}&0.1464\%&0.5076\%\\
		MedianBlur&0.0000\%&\multicolumn{1}{c|}{0.0000\%}&0.0141\%&0.0106\%\\
		Brightness&0.0012\%&\multicolumn{1}{c|}{0.0012\%}&0.0035\%&0.0083\%\\
		Contrast&0.0012\%&\multicolumn{1}{c|}{0.0000\%}&0.0224\%&0.0224\%\\
		Saturation&0.0000\%&\multicolumn{1}{c|}{0.0000\%}&0.0000\%&0.0000\%\\
		Hue&0.0000\%&\multicolumn{1}{c|}{0.0000\%}&0.0000\%&0.0000\%\\
		Dropout&0.0012\%&\multicolumn{1}{c|}{0.0012\%}&0.0024\%&0.0047\%\\
		SaltPepper&0.0059\%&\multicolumn{1}{c|}{0.0083\%}&0.4308\%&0.3778\%\\
		GaussianNoise&0.4273\%&\multicolumn{1}{c|}{0.4403\%}&2.0515\%&2.2238\%\\
		SimSwap&15.8015\%&\multicolumn{1}{c|}{54.1714\%}&20.0802\%&48.3239\%\\
		GANimation&0.0037\%&\multicolumn{1}{c|}{53.9691\%}&0.0012\%&29.5494\%\\
		StarGAN (Male)&0.0331\%&\multicolumn{1}{c|}{58.0914\%}&0.3175\%&45.4544\%\\
		StarGAN (Young) &0.0378\%&\multicolumn{1}{c|}{57.7679\%}&0.3671\%&45.2325\%\\
		StarGAN (BlackHair) &0.0425\%&\multicolumn{1}{c|}{60.0248\%}&0.4355\%&45.3978\%\\
		StarGAN (BlondHair) &0.0354\%&\multicolumn{1}{c|}{56.9795\%}&0.4202\%&45.3872\%\\
		StarGAN (BrownHair) &0.0189\%&\multicolumn{1}{c|}{59.2694\%}&0.2927\%&45.0342\%\\
		\hline
		Average Common&0.0454\%&\multicolumn{1}{c|}{0.0476\%}&0.2856\%&0.3312\%\\
		&$\downarrow$\textbf{0.0394\%}&\multicolumn{1}{c|}{$\downarrow$\textbf{0.0475\%}}&$\uparrow$\textbf{0.2008\%}&$\uparrow$\textbf{0.2961\%}\\
		Average Malicious&2.2818\%&\multicolumn{1}{c|}{57.1819\%}&3.1306\%&43.4828\%\\
		&$\uparrow$\textbf{0.2162\%}&\multicolumn{1}{c|}{$\overline{\quad\quad}$}&$\uparrow$\textbf{1.0650\%}&$\overline{\quad\quad}$\\
		\bottomrule
	\end{tabular}
\end{table}

\subsection{Visual Quality}
We evaluate the visual quality of encoded images from the objective and subjective perspectives. As we can see in Table~\ref{tab:objective}, SepMark has slightly better performance than PIMoG and is superior to MBRS and FaceSigns by a large margin, in terms of PSNR, SSIM, and LPIPS. Moreover, no clear artifacts can be observed by the naked eye from the encoded images, as visualized in Fig.~\ref{fig:subjective}. We speculate that the satisfied performance mainly derives from our stable adversarial training. Although CIN yields impressive objective visual quality, its invertible neural network is not flexible to be compatible with our architecture of deep separable watermarking, and it has non-ideal robustness when encountering quantization noise.

\subsection{Robustness Test}
Primarily, the robust Tracer should have a low BER under arbitrary distortions to trace the trusted source, while the semi-robust Detector should have a low BER under common distortions but a high BER under malicious ones to distinguish the type of the distortion.

Under the common distortions presented in Table~\ref{tab:common_distortion}, the BER of Tracer approaches $0\%$ on average, reaching nearly optimal robustness compared to the baselines in addition to competitive MBRS. Moreover, the BER of Detector also gains reduction compared to CIN, PIMoG, and FaceSigns. By observation from the ``Both'' column, it can be drawn that the separable decoders extract nearly the same message when encountering common distortions. We do not intend to demonstrate that SepMark is definitely superior to baselines in terms of robustness, but rather that its sufficient robustness against common distortions.

Under the malicious distortions presented in Table~\ref{tab:malicious_distortion}, the Tracer also has ideal robustness compared to the baselines, and its BER reaches the lowest on average. As expected, the BER of Detector approaches $50\%$ on average, close to random guesses. Our Detector successfully holds selective fragility to malicious distortions, while the overly robust MBRS, PIMoG, and Tracer are not. It can also be observed that the separable decoders extract mismatched messages when encountering malicious distortions. 

Intriguingly, the above even holds true for the pristine image that is without the embedded message, and we show in Table~\ref{tab:pristine_image} in the Appendix that the two extracted messages are mismatched under malicious distortions, while they are matched to some extent under common distortions. Also in Table~\ref{tab:image_size} in the Appendix, we investigate the observation that the model trained on images of size $128\times 128$ can work on input images of arbitrary size, thanks to the interpolation when fusing the watermark into image features.

Please refer to Table~\ref{tab:cross_dataset_celeba} in the Appendix for the cross-dataset setting on CelebA; it has minor fluctuation in the BER of each distortion compared to that on CelebA-HQ, as well as in the PSNR of encoded images, thereby validating the excellent generalizability. We also show in Table~\ref{tab:cross_dataset_coco} in the Appendix that the model trained on face images can surprisingly work on the natural images in COCO.

\subsection{Ablation Study}
\begin{table}[t]
	\caption{Ablation experiments. One decoder is trained followed by the other. Image size: $128\times 128$.}
	\label{tab:two_stage}
	\small
	\renewcommand\arraystretch{0.9}
	\setlength{\tabcolsep}{4pt}
	\begin{tabular}{lcccc}
		& \multicolumn{2}{c|}{Tracer$\to$Detector} & \multicolumn{2}{c}{Detector$\to$Tracer}\\
		\toprule
		\multirow{2}{*}{Distortion}&\multicolumn{2}{c|}{PSNR $=39.3359$}& \multicolumn{2}{c}{PSNR $=42.6461$}\\
		{}&Tracer&\multicolumn{1}{c|}{Detector}& Tracer & Detector\\
		\midrule
		Identity&0.0000\%&\multicolumn{1}{c|}{0.0000\%}&0.0012\%&0.0000\%\\
		JpegTest&0.6645\%&\multicolumn{1}{c|}{0.5430\%}&2.0267\%&2.4906\%\\
		Resize&0.0295\%&\multicolumn{1}{c|}{0.0212\%}&0.1039\%&0.0720\%\\
		GaussianBlur&0.0012\%&\multicolumn{1}{c|}{0.0000\%}&0.0260\%&0.0189\%\\
		MedianBlur&0.0012\%&\multicolumn{1}{c|}{0.0012\%}&0.0059\%&0.0201\%\\
		Brightness&0.0000\%&\multicolumn{1}{c|}{0.0012\%}&0.0035\%&0.0047\%\\
		Contrast&0.0035\%&\multicolumn{1}{c|}{0.0035\%}&0.0106\%&0.0071\%\\
		Saturation&0.0000\%&\multicolumn{1}{c|}{0.0000\%}&0.0000\%&0.0000\%\\
		Hue&0.0000\%&\multicolumn{1}{c|}{0.0000\%}&0.0000\%&0.0000\%\\
		Dropout&0.0035\%&\multicolumn{1}{c|}{0.0059\%}&0.0047\%&0.0024\%\\
		SaltPepper&0.1333\%&\multicolumn{1}{c|}{0.0024\%}&0.0342\%&0.2998\%\\
		GaussianNoise&1.5557\%&\multicolumn{1}{c|}{1.5321\%}&3.5057\%&4.1218\%\\
		SimSwap&17.9438\%&\multicolumn{1}{c|}{52.1966\%}&29.1678\%&48.5045\%\\
		GANimation&0.0062\%&\multicolumn{1}{c|}{49.2074\%}&0.0198\%&47.3840\%\\
		StarGAN (Male)&0.2491\%&\multicolumn{1}{c|}{54.3366\%}&0.4001\%&43.1256\%\\
		StarGAN (Young)&0.2656\%&\multicolumn{1}{c|}{53.9731\%}&0.4603\%&43.0961\%\\
		StarGAN (BlackHair)&0.3376\%&\multicolumn{1}{c|}{53.2153\%}&0.5453\%&43.0146\%\\
		StarGAN (BlondHair)&0.4084\%&\multicolumn{1}{c|}{54.2292\%}&0.5382\%&42.6039\%\\
		StarGAN (BrownHair)& 0.2077\%&\multicolumn{1}{c|}{53.5694\%}&0.4344\%&42.9792\%\\
		\hline
		Average Common&0.1994\%&\multicolumn{1}{c|}{0.1759\%}&0.4769\%&0.5865\%\\
		&$\uparrow$\textbf{0.1146\%}&\multicolumn{1}{c|}{$\uparrow$\textbf{0.0808\%}}&$\uparrow$\textbf{0.3921\%}&$\uparrow$\textbf{0.4914\%}\\
		Average Malicious&2.7741\%&\multicolumn{1}{c|}{52.9611\%}&4.5094\%&44.3868\%\\
		&$\uparrow$\textbf{0.7085\%}&\multicolumn{1}{c|}{$\overline{\quad\quad}$}&$\uparrow$\textbf{2.4438\%}&$\overline{\quad\quad}$\\
		\bottomrule
	\end{tabular}
\end{table}

We declare that the lack of any component in our architecture may result in unsatisfied training. For the ablation in an effective setup, we examine the range of embedded message, i.e., the hyper-parameter $\alpha$. Accordingly, as depicted in Table~\ref{tab:message_range}, a smaller $\alpha$ can achieve higher visual quality but with reduced robustness, while a larger $\alpha$ can achieve stronger robustness but with decreased visual quality. Recall that $\alpha$ bridges the trade-off between visual quality and robustness. By the way, the bold values in the table mean the performance gains compared to the original counterparts (basically refers to the values in Tables~\ref{tab:common_distortion} and \ref{tab:malicious_distortion}). The omitted term indicates the BER comparison under malicious distortion, as they are close to $50\%$ (random guesses), but higher is not necessarily better.

It is worth mentioning that we also try the two-stage learning, where one decoder is jointly trained with the left components, followed by training with only the other decoder for an additional $100$ epochs. From Table~\ref{tab:two_stage} we can see that training the robust decoder first and then training the semi-robust decoder, or conversely, is inferior to the end-to-end training of SepMark in terms of robustness (see Tables~\ref{tab:common_distortion} and \ref{tab:malicious_distortion}). One hypothesis is that this is likely due to the unbalanced trade-off with visual quality. Furthermore, we conduct the toy example where MBRS first embeds a robust watermark into the image and then uses FaceSigns to embed a second fragile watermark and vice versa, causing the failure of robust decoding and the deteriorated visual quality, as seen in Table~\ref{tab:toy_example} in the Appendix. All of these again demonstrate the effectiveness of the proposed SepMark.

\section{Conclusion}
In this paper, we propose the first end-to-end learning architecture for deep separable watermarking, dubbed SepMark, and naturally apply it to proactive Deepfake forensics, realizing unified source tracing and Deepfake detection. Thanks to the separable decoders that extract the embedded watermark with different levels of robustness, the robust decoder resists against numerous distortions while the semi-robust one is selectively sensitive to malicious distortions. Experimental results show that SepMark has made an encouraging start on typical Deepfakes, to provide preventive measures and more reliable evidence for trustworthy faces. Further work involves the more general modeling of Deepfake distortions that can generalize the watermarking to unseen Deepfakes.

\begin{acks}
This work is supported by National Natural Science Foundation of China
(Grant Nos. U22A2030, 61972142), National Key R\&D Program of China
(Grant No. 2022YFB3103500).
\end{acks}

\bibliographystyle{ACM-Reference-Format}
\balance
\bibliography{sample-base}

\clearpage
\appendix
\section{Appendix}
\subsection{Additional Visual Examples}\label{sec:additional_visual_examples}
\begin{figure}[h]
	\centering
	\subfloat[Image size: $128\times 128$]{\includegraphics[width=0.4\columnwidth]{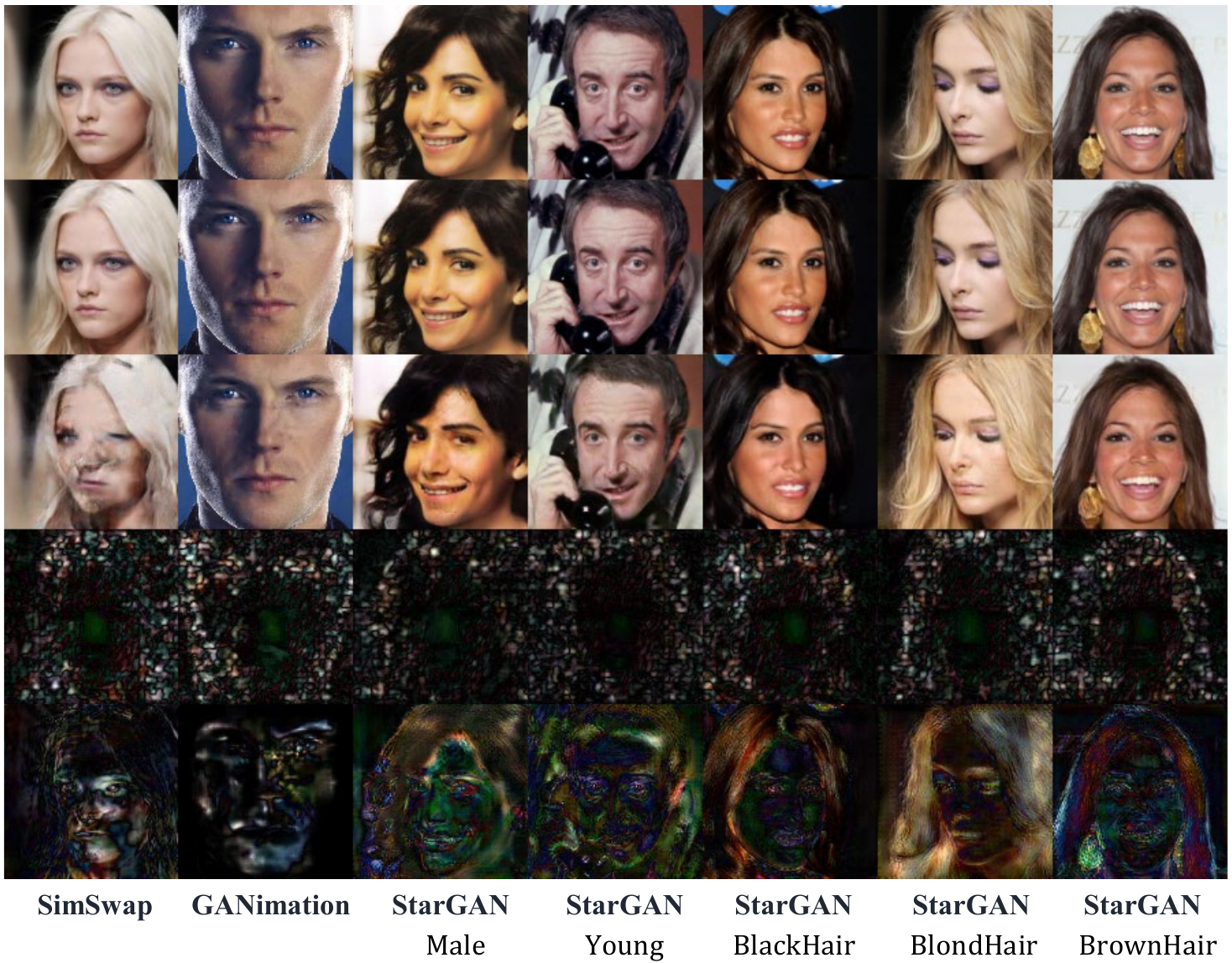}}\hspace{10pt}
	\subfloat[Image size: $256\times 256$]{\includegraphics[width=0.4\columnwidth]{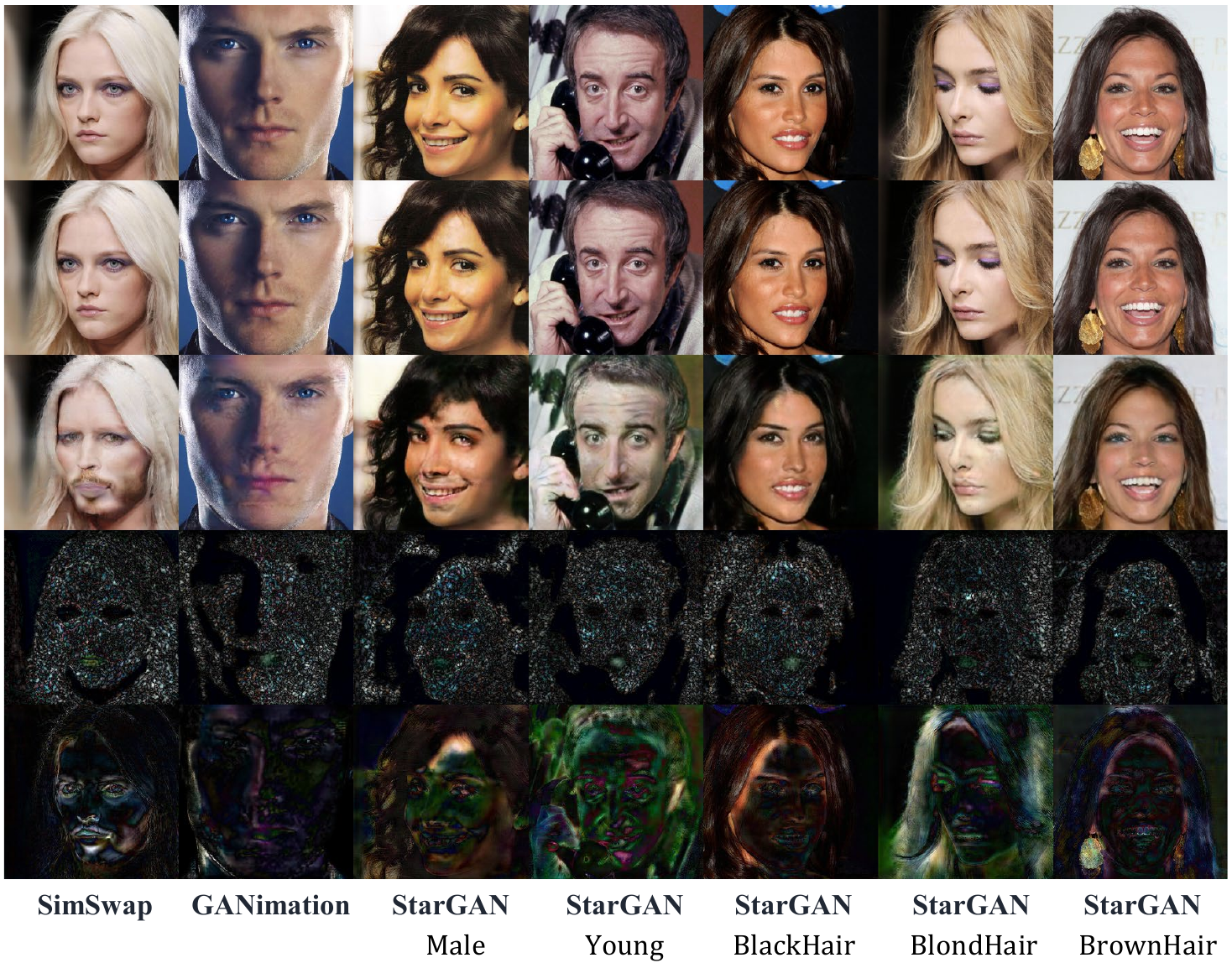}}
	\caption{More malicious distortions. From top to bottom are the cover image $I_{co}$, the encoded image $I_{en}$, the noised image $I_{no}$, the residual signal of $\mathcal{N}(|I_{co}-I_{en}|)$, and $\mathcal{N}(|I_{no}-I_{en}|)$.}\label{fig:malicious_distortions}
	\Description{More malicious distortions.}
\end{figure}

\subsection{Additional Experimental Results}

\begin{table}[b]
	\caption{More investigations on the pristine image $I_{co}$. ``Both'' means the BER between $M_{tr}$ and $M_{de}$.}
	\label{tab:pristine_image}
	\scriptsize
	\renewcommand\arraystretch{0.8}
	\setlength{\tabcolsep}{14.5pt}
	\begin{tabular}{lcc}
		& \multicolumn{1}{c|}{$128\times 128$} & \multicolumn{1}{c}{$256\times 256$}\\
		\toprule
		\multirow{2}{*}{Distortion}&\multicolumn{1}{c|}{PSNR $= +\infty$}& \multicolumn{1}{c}{PSNR $= +\infty$}\\
		{}&\multicolumn{1}{c|}{Both}& Both\\
		\midrule
		Identity&\multicolumn{1}{c|}{18.3534\%}&15.8034\%\\
		JpegTest&\multicolumn{1}{c|}{14.3272\%}&16.7296\%\\
		Resize&\multicolumn{1}{c|}{17.1341\%}&15.4153\%\\
		GaussianBlur&\multicolumn{1}{c|}{18.0217\%}&17.9925\%\\
		MedianBlur&\multicolumn{1}{c|}{17.6121\%}&17.0536\%\\
		Brightness&\multicolumn{1}{c|}{18.7901\%}&16.7889\%\\
		Contrast&\multicolumn{1}{c|}{18.6202\%}&16.8386\%\\
		Saturation&\multicolumn{1}{c|}{18.3392\%}&15.9481\%\\
		Hue&\multicolumn{1}{c|}{18.2649\%}&15.8679\%\\
		Dropout&\multicolumn{1}{c|}{18.3534\%}&15.8034\%\\
		SaltPepper&\multicolumn{1}{c|}{27.8199\%}&23.0745\%\\
		GaussianNoise&\multicolumn{1}{c|}{16.0966\%}&18.7992\%\\
		SimSwap&\multicolumn{1}{c|}{50.5406\%}&47.4623\%\\
		GANimation&\multicolumn{1}{c|}{48.6370\%}&49.8244\%\\
		StarGAN (Male)&\multicolumn{1}{c|}{50.4049\%}&49.0832\%\\
		StarGAN (Young) &\multicolumn{1}{c|}{50.6209\%}&49.2378\%\\
		StarGAN (BlackHair) &\multicolumn{1}{c|}{49.5609\%}&48.9125\%\\
		StarGAN (BlondHair) &\multicolumn{1}{c|}{48.7028\%}&48.9916\%\\
		StarGAN (BrownHair) &\multicolumn{1}{c|}{50.1771\%}&49.0907\%\\
		\hline
		Average Common&\multicolumn{1}{c|}{18.4777\%}&17.1763\%\\
		Average Malicious&\multicolumn{1}{c|}{49.8063\%}&48.9432\%\\
		\bottomrule
	\end{tabular}
\end{table}

\begin{table}[b]
	\caption{More investigations on $128\times 128$ trained model across arbitrary input size of image $I_{co}$.}
	\label{tab:image_size}
	\scriptsize
	\renewcommand\arraystretch{0.8}
	\setlength{\tabcolsep}{3.5pt}
	\begin{tabular}{lcccc}
		& \multicolumn{2}{c|}{$256\times 256$} & \multicolumn{2}{c}{$512\times 512$}\\
		\toprule
		\multirow{2}{*}{Distortion}&\multicolumn{2}{c|}{PSNR $=35.4984$}& \multicolumn{2}{c}{PSNR $=35.2925$}\\
		{}&Tracer&\multicolumn{1}{c|}{Detector}& Tracer & Detector\\
		\midrule
		Identity&0.0000\%&\multicolumn{1}{c|}{5.2467\%}&0.0000\%&25.1287\%\\
		JpegTest&0.1735\%&\multicolumn{1}{c|}{0.9679\%}&0.1062\%&0.7177\%\\
		Resize&0.0508\%&\multicolumn{1}{c|}{2.9592\%}&0.0165\%&18.6532\%\\
		GaussianBlur&0.0106\%&\multicolumn{1}{c|}{11.0127\%}&0.0035\%&33.1704\%\\
		MedianBlur&0.0071\%&\multicolumn{1}{c|}{9.0061\%}&0.0035\%&30.9301\%\\
		Brightness&0.0153\%&\multicolumn{1}{c|}{5.8735\%}&0.0165\%&25.0118\%\\
		Contrast&0.0177\%&\multicolumn{1}{c|}{6.2028\%}&0.0106\%&25.3234\%\\
		Saturation&0.0000\%&\multicolumn{1}{c|}{5.4958\%}&0.0000\%&25.3966\%\\
		Hue&0.0000\%&\multicolumn{1}{c|}{5.5902\%}&0.0000\%&26.1036\%\\
		Dropout&0.0165\%&\multicolumn{1}{c|}{7.6546\%}&0.0224\%&26.1674\%\\
		SaltPepper&0.1558\%&\multicolumn{1}{c|}{2.0491\%}&0.2325\%&5.9832\%\\
		GaussianNoise&0.5323\%&\multicolumn{1}{c|}{3.9341\%}&0.3895\%&8.1161\%\\
		SimSwap&2.5484\%&\multicolumn{1}{c|}{49.9115\%}&1.9677\%&49.1407\%\\
		GANimation&0.0679\%&\multicolumn{1}{c|}{46.8494\%}&0.1136\%&44.9160\%\\
		StarGAN (Male)&0.2668\%&\multicolumn{1}{c|}{50.2101\%}&0.0873\%&49.8430\%\\
		StarGAN (Young) &0.2821\%&\multicolumn{1}{c|}{50.0390\%}&0.0814\%&49.4653\%\\
		StarGAN (BlackHair) &0.2986\%&\multicolumn{1}{c|}{50.6232\%}&0.0755\%&50.8263\%\\
		StarGAN (BlondHair) &0.3553\%&\multicolumn{1}{c|}{49.5574\%}&0.0921\%&48.2177\%\\
		StarGAN (BrownHair) &0.2644\%&\multicolumn{1}{c|}{50.3057\%}&0.0767\%&50.3836\%\\
		\hline
		Average Common&0.0816\%&\multicolumn{1}{c|}{5.4994\%}&0.0668\%&20.8919\%\\
		&$\downarrow$\textbf{0.0032\%}&\multicolumn{1}{c|}{$\uparrow$\textbf{5.4043\%}}&$\downarrow$\textbf{0.0180\%}&$\uparrow$\textbf{20.7968\%}\\
		Average Malicious&0.5834\%&\multicolumn{1}{c|}{49.6423\%}&0.3563\%&48.9704\%\\
		&$\downarrow$\textbf{1.4822\%}&\multicolumn{1}{c|}{$\overline{\quad\quad}$}&$\downarrow$\textbf{1.7093\%}&$\overline{\quad\quad}$\\
		\bottomrule
	\end{tabular}
\end{table}

\begin{table}[b]
	\caption{Robustness test for the image $I_{no}$ under common and malicious distortions on cross-dataset CelebA.}
	\label{tab:cross_dataset_celeba}
	\scriptsize
	\renewcommand\arraystretch{0.97}
	\setlength{\tabcolsep}{4pt}
	\begin{tabular}{lcccccc}
		& \multicolumn{2}{c|}{$128\times 128$} & \multicolumn{2}{c}{$256\times 256$}\\
		\toprule
		\multirow{2}{*}{Distortion}&\multicolumn{2}{c|}{PSNR $=37.8865$}&\multicolumn{2}{c}{PSNR $=38.7051$}\\
		&{Tracer}&\multicolumn{1}{c|}{Detector}&{Tracer}&{Detector}\\
		\midrule
		Identity &0.0000\%&\multicolumn{1}{c|}{0.0417\%}&0.0063\%&0.0002\%\\
		JpegTest &0.2233\%&\multicolumn{1}{c|}{0.2934\%}&0.0247\%&0.0194\%\\
		Resize &0.0134\%&\multicolumn{1}{c|}{0.0272\%}&0.0001\%&0.0002\%\\
		GaussianBlur &0.0058\%&\multicolumn{1}{c|}{0.0416\%}&0.0004\%&0.0022\%\\
		MedianBlur &0.0037\%&\multicolumn{1}{c|}{0.0730\%}&0.0003\%&0.0002\%\\
		Brightness &0.0094\%&\multicolumn{1}{c|}{0.0579\%}&0.0160\%&0.0016\%\\
		Contrast &0.0068\%&\multicolumn{1}{c|}{0.0698\%}&0.0034\%&0.0002\%\\
		Saturation &0.0000\%&\multicolumn{1}{c|}{0.0519\%}&0.0056\%&0.0002\%\\
		Hue &0.0000\%&\multicolumn{1}{c|}{0.0401\%}&0.0070\%&0.0002\%\\
		Dropout &0.0020\%&\multicolumn{1}{c|}{0.0625\%}&0.4647\%&0.0004\%\\
		SaltPepper &0.0406\%&\multicolumn{1}{c|}{0.0242\%}&0.0017\%&0.0005\%\\
		GaussianNoise &0.7461\%&\multicolumn{1}{c|}{0.8952\%}&0.1594\%&0.1708\%\\
		SimSwap &12.2428\%&\multicolumn{1}{c|}{50.7506\%}&10.2216\%&34.2429\%\\
		GANimation &0.0007\%&\multicolumn{1}{c|}{39.6265\%}&0.0001\%&45.5291\%\\
		StarGAN (Male) &0.1209\%&\multicolumn{1}{c|}{50.8483\%}&0.0019\%&45.5910\%\\
		StarGAN (Young) &0.1358\%&\multicolumn{1}{c|}{50.6294\%}&0.0025\%&45.6882\%\\
		StarGAN (BlackHair) &0.1479\%&\multicolumn{1}{c|}{48.0443\%}&0.0021\%&46.1099\%\\
		StarGAN (BlondHair) &0.1840\%&\multicolumn{1}{c|}{46.7939\%}&0.0020\%&44.6107\%\\
		StarGAN (BrownHair) &0.1196\%&\multicolumn{1}{c|}{50.8680\%}&0.0014\%&45.9944\%\\
		\hline
		Average Common&0.0876\%&\multicolumn{1}{c|}{0.1399\%}&0.0575\%&0.0163\%\\
		&$\uparrow$\textbf{0.0028\%}&\multicolumn{1}{c|}{$\uparrow$\textbf{0.0448\%}}&$\uparrow$\textbf{0.0514\%}&$\uparrow$\textbf{0.0080\%}\\
		Average Malicious&1.8502\%&\multicolumn{1}{c|}{48.2230\%}&1.4617\%&43.9666\%\\
		&$\downarrow$\textbf{0.2154\%}&\multicolumn{1}{c|}{$\overline{\quad\quad}$}&$\uparrow$\textbf{0.3301\%}&$\overline{\quad\quad}$\\
		\bottomrule
	\end{tabular}
\end{table}

\begin{table}[b]
	\caption{Robustness test for the common distorted image $I_{no}$ on cross-dataset COCO.}
	\label{tab:cross_dataset_coco}
	\scriptsize
	\renewcommand\arraystretch{0.97}
	\setlength{\tabcolsep}{4pt}
	\begin{tabular}{lcccccc}
		& \multicolumn{2}{c|}{$128\times 128$} & \multicolumn{2}{c}{$256\times 256$}\\
		\toprule
		\multirow{2}{*}{Distortion}&\multicolumn{2}{c|}{PSNR $=35.1500$}&\multicolumn{2}{c}{PSNR $=35.5506$}\\
		&{Tracer}&\multicolumn{1}{c|}{Detector}&{Tracer}&{Detector}\\
		\midrule
		Identity &0.2100\%&\multicolumn{1}{c|}{1.4440\%}&0.0420\%&0.0475\%\\
		JpegTest &1.6027\%&\multicolumn{1}{c|}{3.6627\%}&0.1691\%&0.1439\%\\
		Resize &3.9873\%&\multicolumn{1}{c|}{2.8900\%}&0.2661\%&0.2502\%\\
		GaussianBlur &1.1447\%&\multicolumn{1}{c|}{4.0467\%}&0.0698\%&0.1478\%\\
		MedianBlur &0.7460\%&\multicolumn{1}{c|}{3.4340\%}&0.1005\%&0.0878\%\\
		Brightness &0.4560\%&\multicolumn{1}{c|}{1.9213\%}&0.1745\%&0.1500\%\\
		Contrast &0.3860\%&\multicolumn{1}{c|}{1.7367\%}&0.0756\%&0.0639\%\\
		Saturation &0.2333\%&\multicolumn{1}{c|}{1.5240\%}&0.0438\%&0.0567\%\\
		Hue &0.2293\%&\multicolumn{1}{c|}{1.6453\%}&0.0428\%&0.0495\%\\
		Dropout &0.8800\%&\multicolumn{1}{c|}{2.8593\%}&0.2595\%&0.1906\%\\
		SaltPepper &1.8013\%&\multicolumn{1}{c|}{1.9353\%}&0.1944\%&0.1355\%\\
		GaussianNoise &1.0473\%&\multicolumn{1}{c|}{2.2393\%}&0.2186\%&0.2402\%\\
		\hline
		Average Common&1.0603\%&\multicolumn{1}{c|}{2.4449\%}&0.1381\%&0.1303\%\\
		&$\uparrow$\textbf{0.9755\%}&\multicolumn{1}{c|}{$\uparrow$\textbf{2.3498\%}}&$\uparrow$\textbf{0.1320\%}&$\uparrow$\textbf{0.1220\%}\\
		\bottomrule
	\end{tabular}
\end{table}

\begin{table}[b]
	\caption{Toy example. One watermark is embedded followed by the other. Image size: $256\times 256$.}
	\label{tab:toy_example}
	\scriptsize
	\renewcommand\arraystretch{0.97}
	\setlength{\tabcolsep}{3pt}
	\begin{tabular}{lcccc}
		& \multicolumn{2}{c|}{MBRS$\to$FaceSigns} & \multicolumn{2}{c}{FaceSigns$\to$MBRS}\\
		\toprule
		\multirow{2}{*}{Distortion}&\multicolumn{2}{c|}{PSNR $=32.1240$}& \multicolumn{2}{c}{PSNR $=32.0076$}\\
		{}&MBRS&\multicolumn{1}{c|}{FaceSigns}& MBRS & FaceSigns\\
		\midrule
		Identity&0.3378\%&\multicolumn{1}{c|}{0.0133\%}&0.0000\%&0.0340\%\\
		JpegTest&5.0135\%&\multicolumn{1}{c|}{0.8648\%}&0.1546\%&1.5589\%\\
		Resize&5.7166\%&\multicolumn{1}{c|}{1.1027\%}&1.6975\%&1.5783\%\\
		GaussianBlur&6.9225\%&\multicolumn{1}{c|}{0.1328\%}&3.6559\%&0.1676\%\\
		MedianBlur&5.1384\%&\multicolumn{1}{c|}{0.1151\%}&2.3408\%&1.5160\%\\
		Brightness&1.7552\%&\multicolumn{1}{c|}{0.0764\%}&1.0235\%&0.0943\%\\
		Contrast&1.7989\%&\multicolumn{1}{c|}{0.0318\%}&0.9658\%&0.0515\%\\
		Saturation&0.3329\%&\multicolumn{1}{c|}{0.7660\%}&0.0000\%&0.8219\%\\
		Hue&0.9683\%&\multicolumn{1}{c|}{8.0228\%}&0.0296\%&9.3728\%\\
		Dropout&4.5914\%&\multicolumn{1}{c|}{17.6177\%}&38.8410\%&5.9916\%\\
		SaltPepper&40.2750\%&\multicolumn{1}{c|}{12.4776\%}&32.2704\%&13.3133\%\\
		GaussianNoise&25.7147\%&\multicolumn{1}{c|}{0.8963\%}&12.5065\%&1.2598\%\\
		SimSwap&46.5361\%&\multicolumn{1}{c|}{49.8661\%}&45.8514\%&49.8559\%\\
		GANimation&2.7499\%&\multicolumn{1}{c|}{45.2734\%}&0.2568\%&45.5127\%\\
		StarGAN (Male)&11.8858\%&\multicolumn{1}{c|}{50.1740\%}&7.7480\%&50.2683\%\\
		StarGAN (Young)&11.7510\%&\multicolumn{1}{c|}{50.3737\%}&7.8313\%&50.4302\%\\
		StarGAN (BlackHair)&13.0742\%&\multicolumn{1}{c|}{50.1859\%}&8.9936\%&50.2606\%\\
		StarGAN (BlondHair)&10.3177\%&\multicolumn{1}{c|}{50.9066\%}&6.8103\%&51.1755\%\\
		StarGAN (BrownHair)&11.0078\%&\multicolumn{1}{c|}{50.7384\%}&7.1055\%&50.9381\%\\
		\hline
		Average Common&8.2138\%&\multicolumn{1}{c|}{3.5098\%}&7.7905\%&2.9800\%\\
		&$\uparrow$\textbf{8.2077\%}&\multicolumn{1}{c|}{$\uparrow$\textbf{3.5015\%}}&$\uparrow$\textbf{7.7844\%}&$\uparrow$\textbf{2.9717\%}\\
		Average Malicious&15.3318\%&\multicolumn{1}{c|}{49.6454\%}&12.0853\%&49.7773\%\\
		&$\uparrow$\textbf{14.2002\%}&\multicolumn{1}{c|}{$\overline{\quad\quad}$}&$\uparrow$\textbf{10.9537\%}&$\overline{\quad\quad}$\\
		\bottomrule
	\end{tabular}
\end{table}

%%\if
\clearpage
For each figure in the Appendix, from top to bottom are the cover image $I_{co}$, the encoded image $I_{en}$, the noised image $I_{no}$, the residual signal of $\mathcal{N}(|I_{co}-I_{en}|)$, and $\mathcal{N}(|I_{no}-I_{en}|)$.

\begin{figure}[h!]
	\centering
	\includegraphics[width=0.74\linewidth]{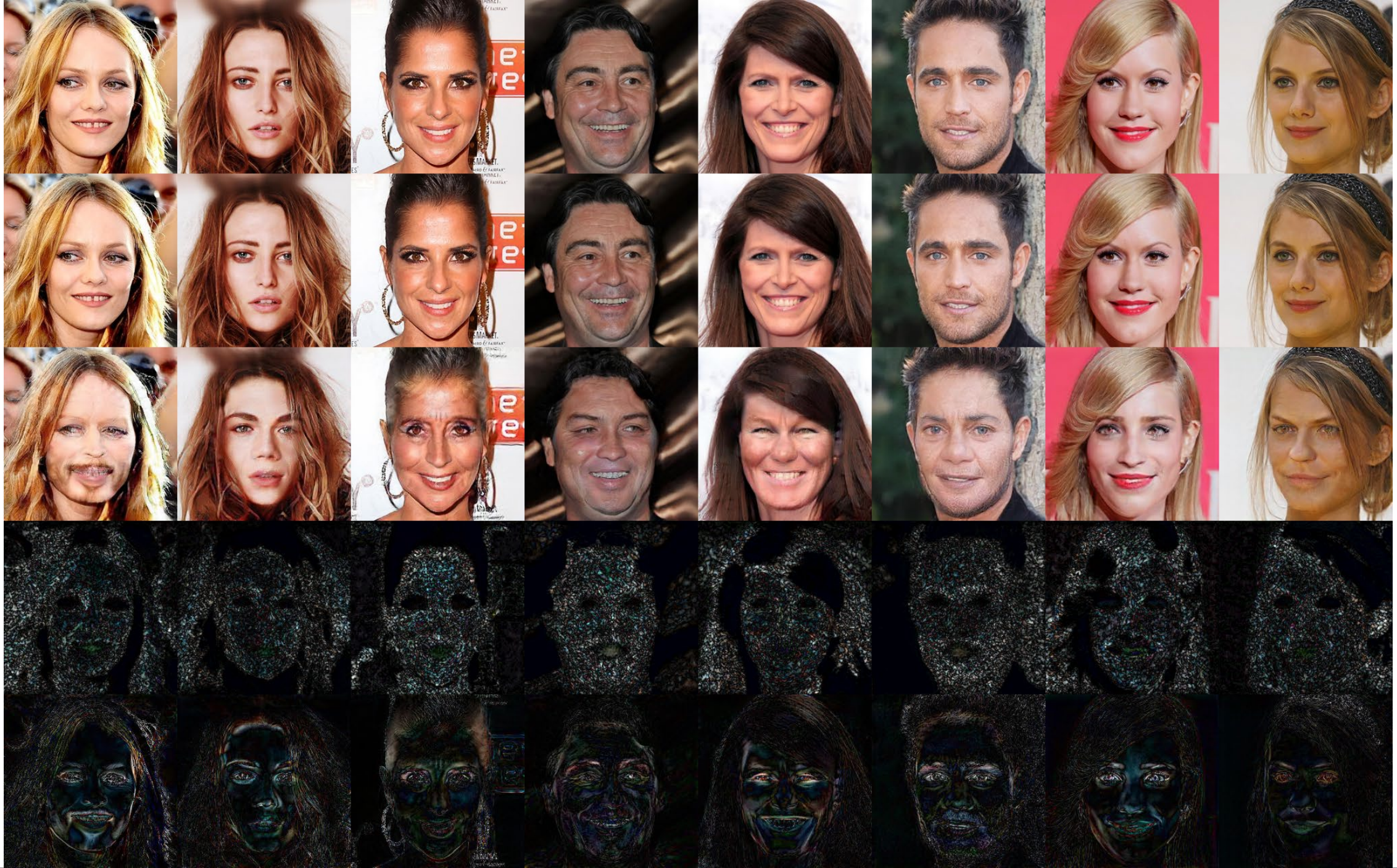}
	\caption{SimSwap. Dataset: CelebA-HQ. Image size: $256\times 256$.}
	\Description{SimSwap.}
\end{figure}
\begin{figure}[h!]
	\centering
	\includegraphics[width=0.74\linewidth]{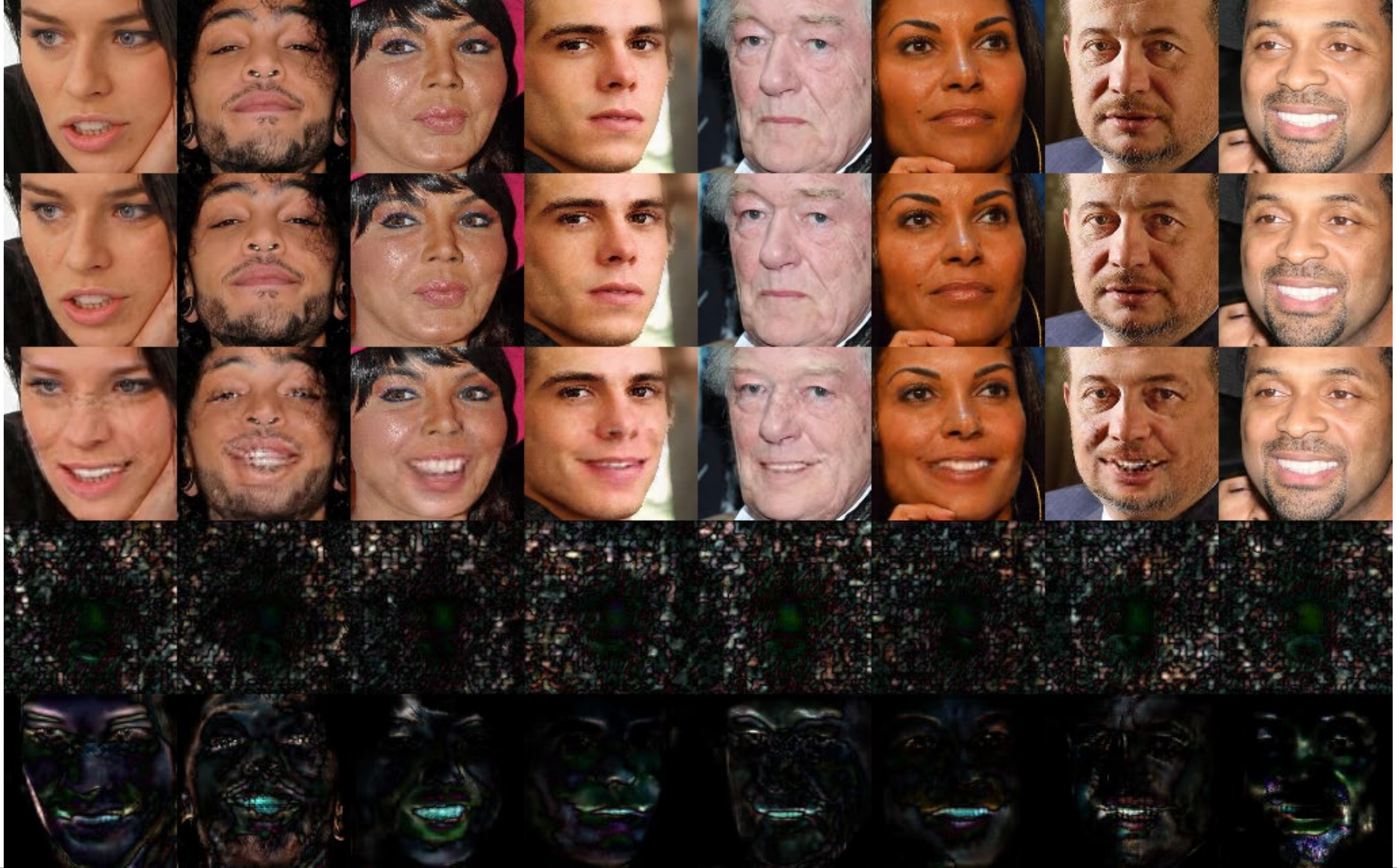}
	\caption{GANimation: the face images are processed by OpenFace only when testing, to match the requirements of the input. Dataset: CelebA-HQ. Image size: $128\times 128$.}
	\Description{GANimation.}
\end{figure}
\begin{figure}[h!]
	\centering
	\includegraphics[width=0.74\linewidth]{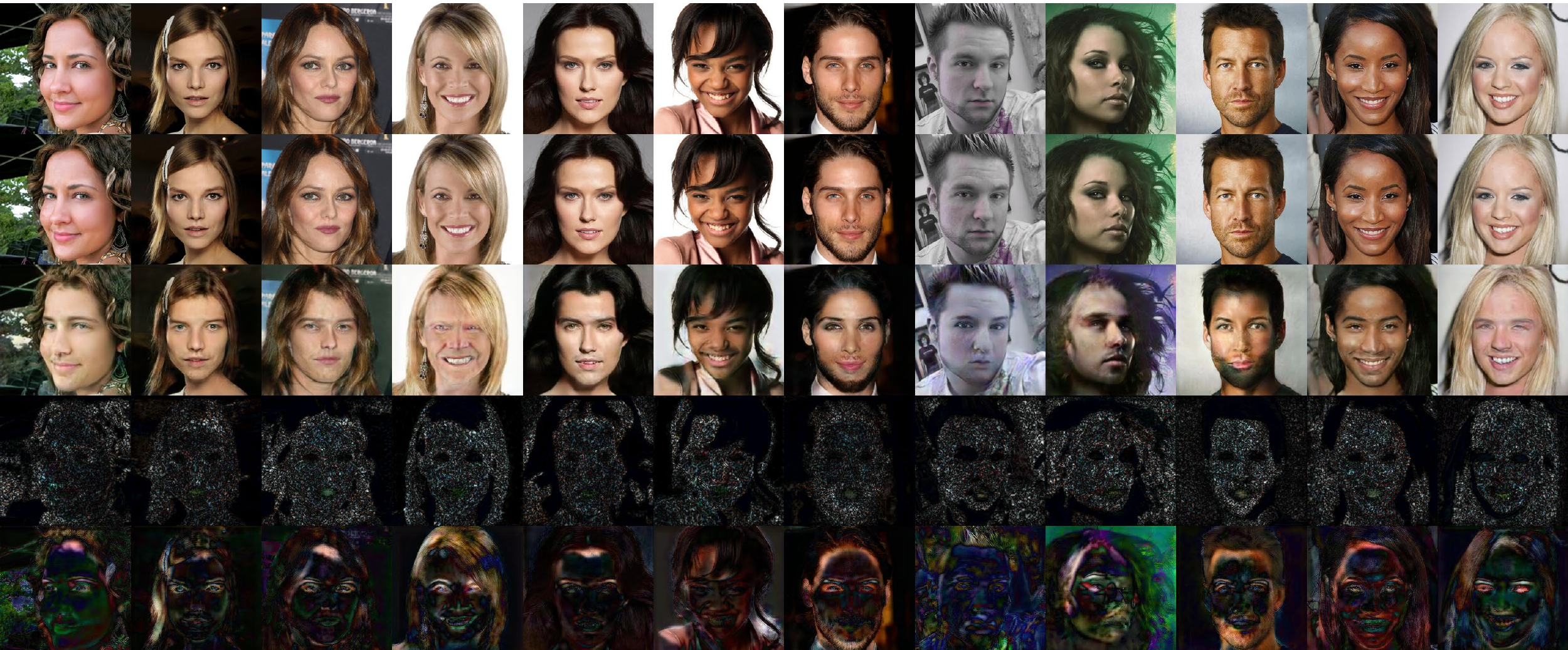}
	\caption{StarGAN (Male). Dataset: CelebA-HQ. Image size: $256\times 256$.}
	\Description{StarGAN (Male).}
\end{figure}
\begin{figure}[b]
	\centering
	\includegraphics[width=0.74\linewidth]{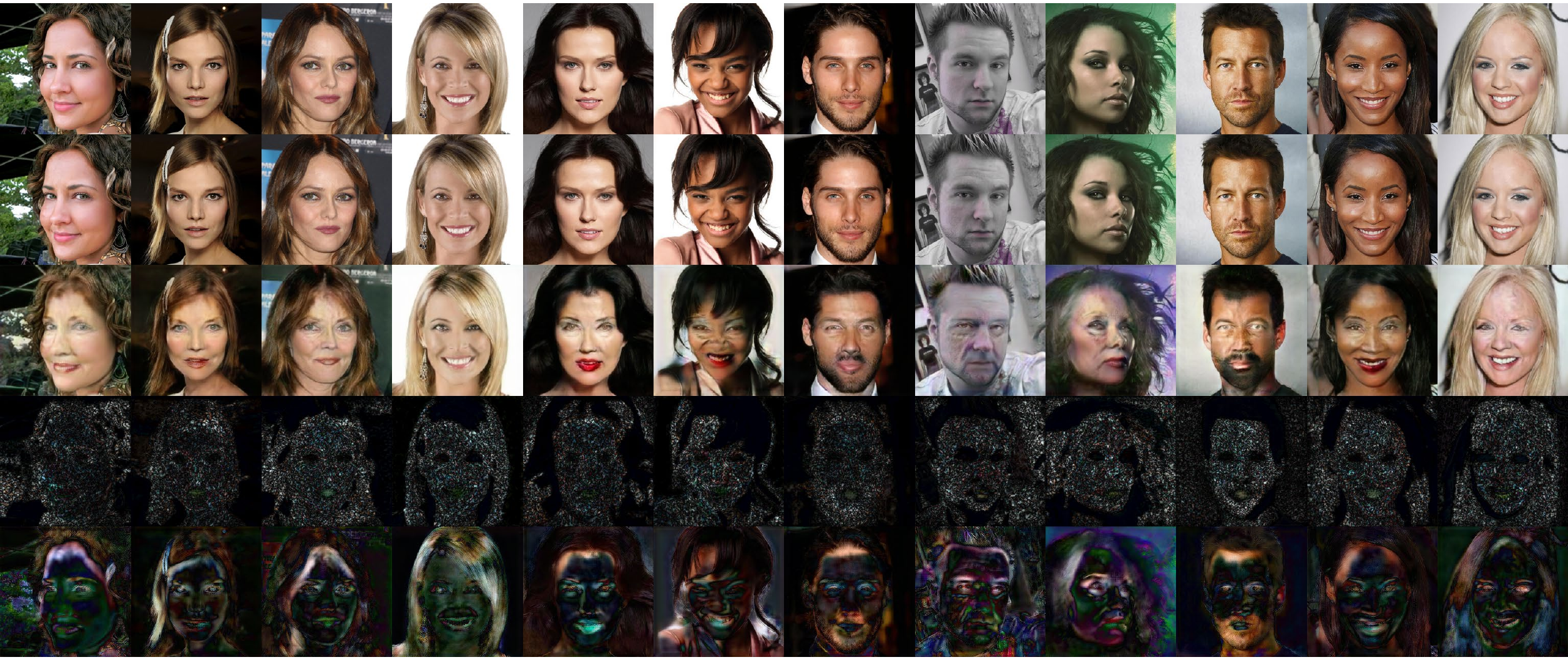}
	\caption{StarGAN (Young). Dataset: CelebA-HQ. Image size: $256\times 256$.}
	\Description{StarGAN (Young).}
\end{figure}
\begin{figure}[b]
	\centering
	\includegraphics[width=0.7\linewidth]{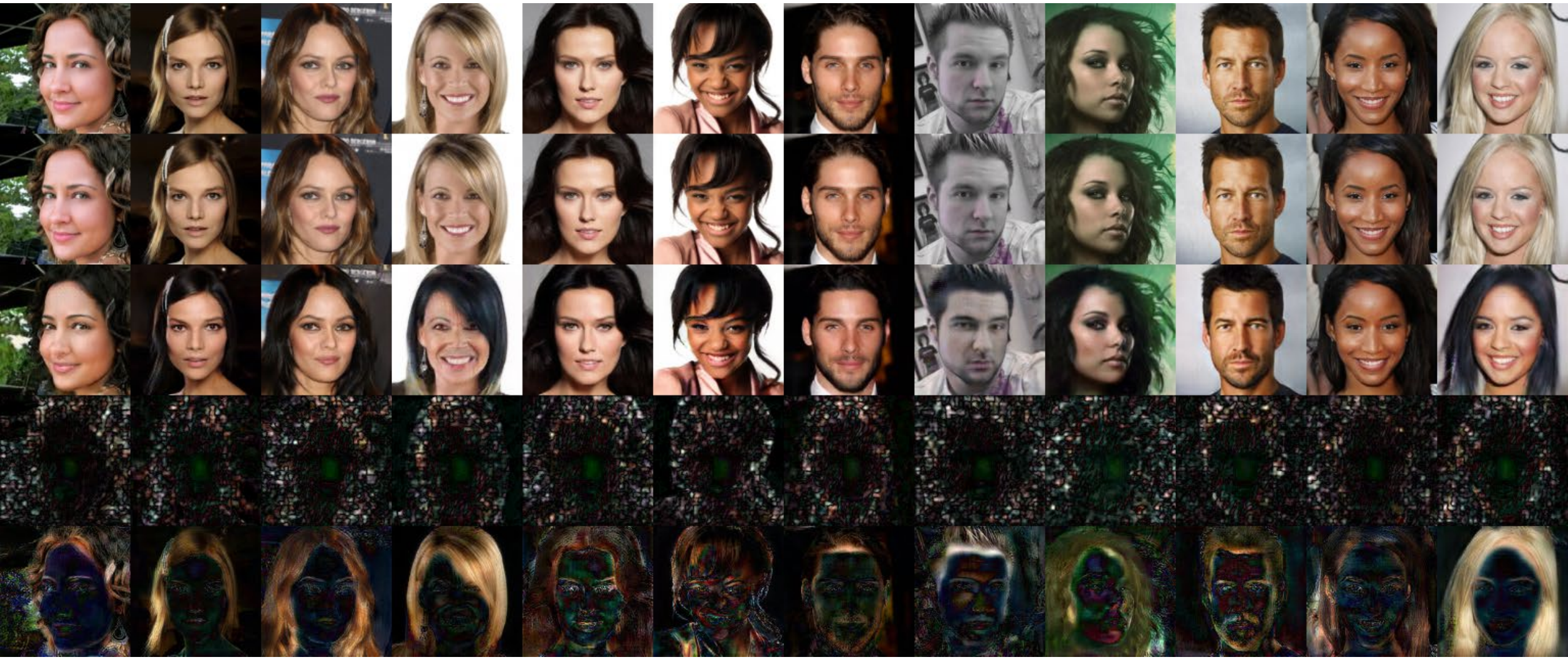}
	\caption{StarGAN (BlackHair). Dataset: CelebA-HQ. Image size: $128\times 128$.}
	\Description{StarGAN (BlackHair).}
\end{figure}
\begin{figure}[b]
	\centering
	\includegraphics[width=0.7\linewidth]{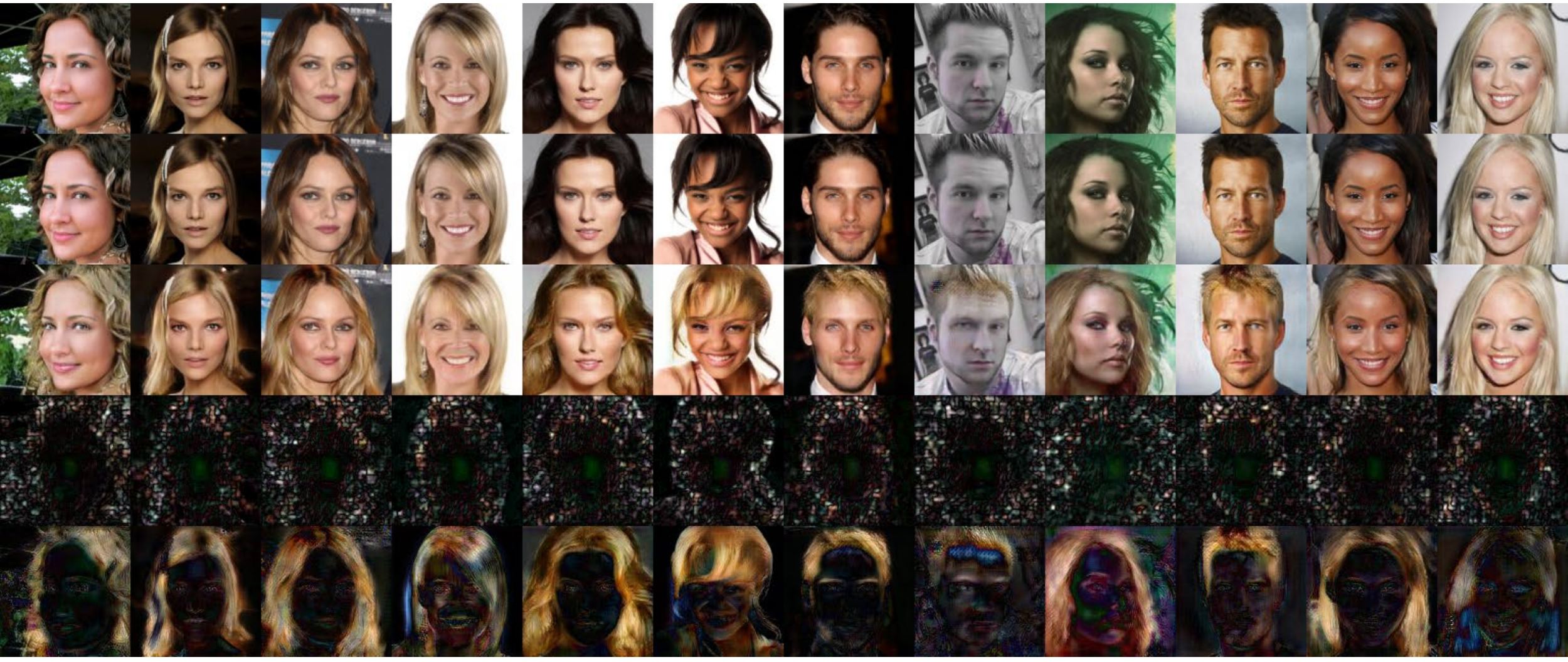}
	\caption{StarGAN (BlondHair). Dataset: CelebA-HQ. Image size: $128\times 128$.}
	\Description{StarGAN (BlondHair).}
\end{figure}
\begin{figure}[b]
	\centering
	\includegraphics[width=0.7\linewidth]{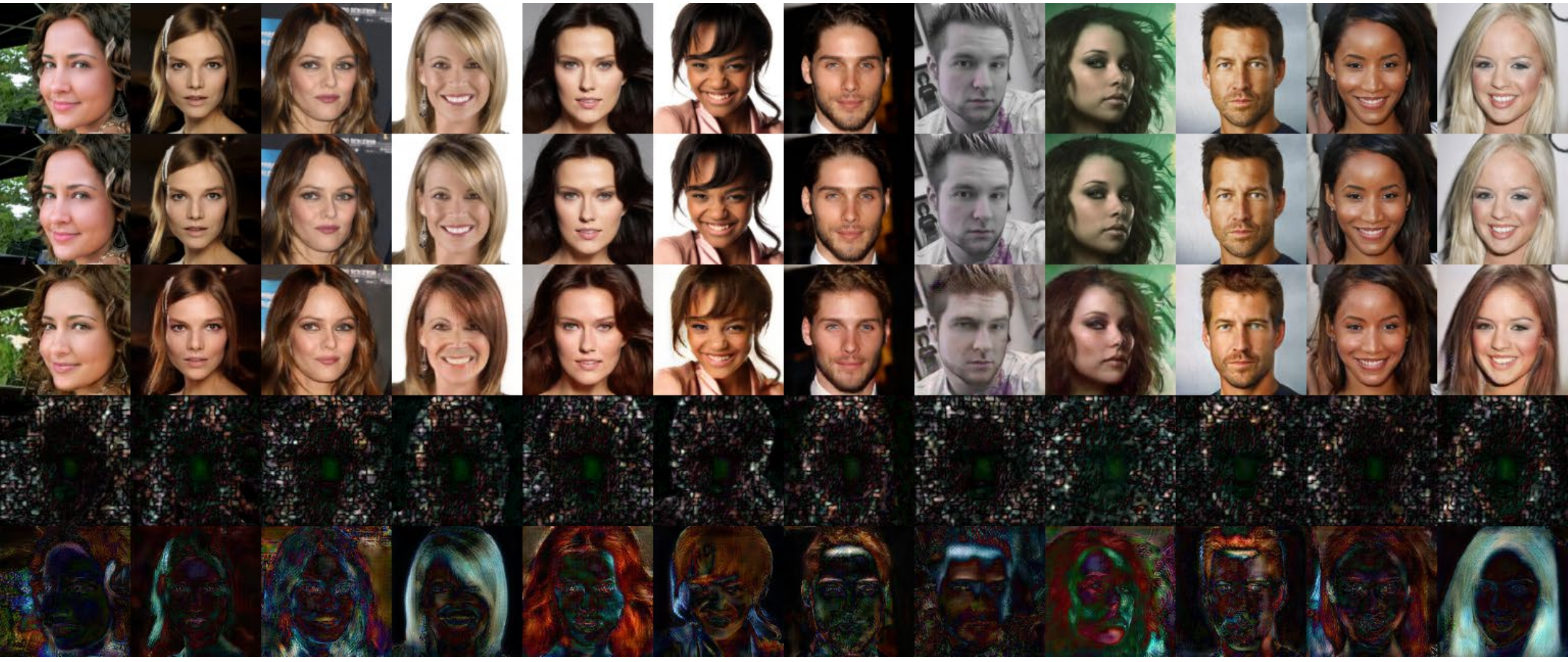}
	\caption{StarGAN (BrownHair). Dataset: CelebA-HQ. Image size: $128\times 128$.}
	\Description{StarGAN (BrownHair).}
\end{figure}
\begin{figure}[b!]
	\centering
	\includegraphics[width=0.7\linewidth]{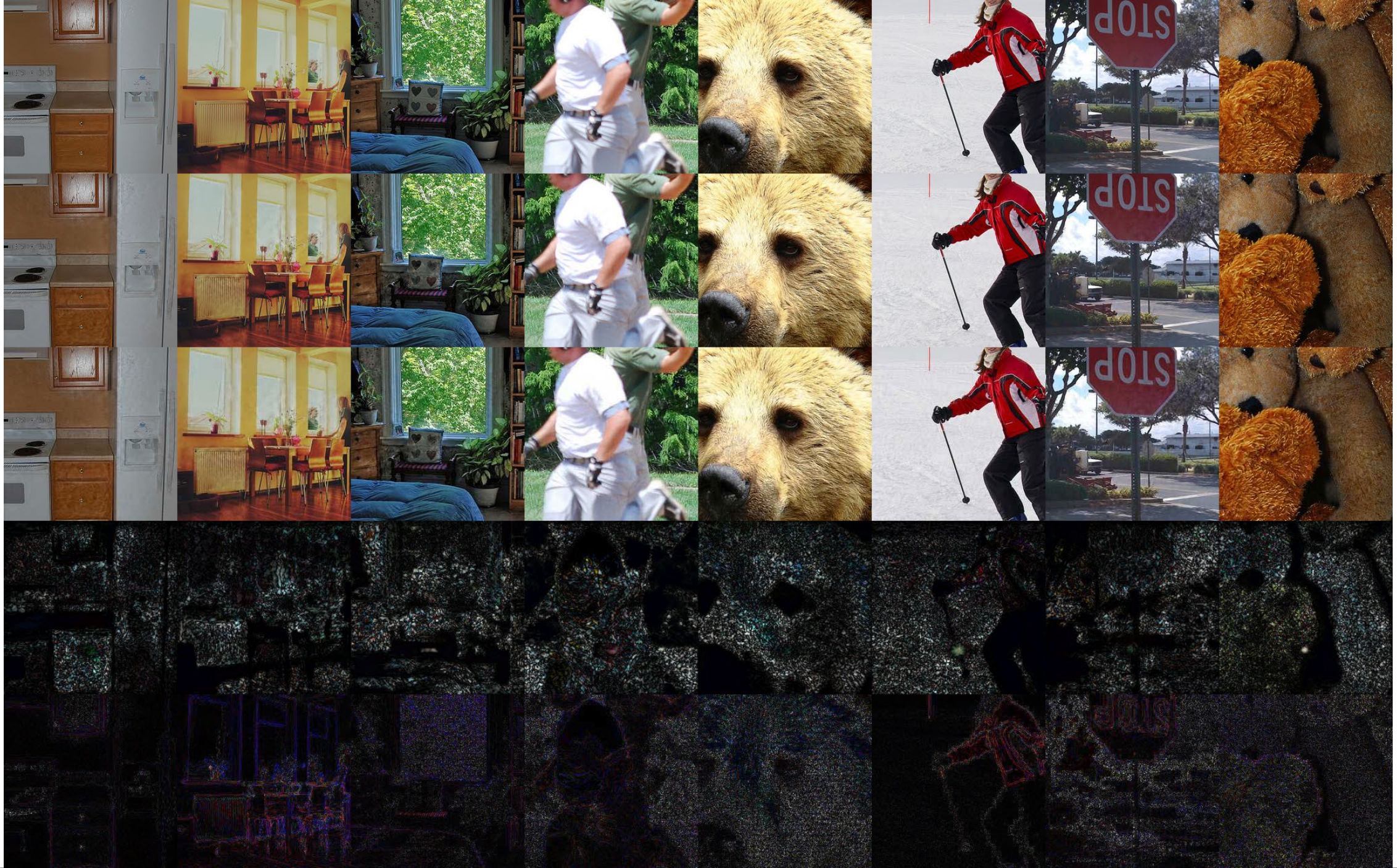}
	\caption{JpegTest. Dataset: COCO. Image size: $256\times 256$.}
	\Description{JpegTest.}
\end{figure}
\begin{figure}[b!]
	\centering
	\includegraphics[width=0.7\linewidth]{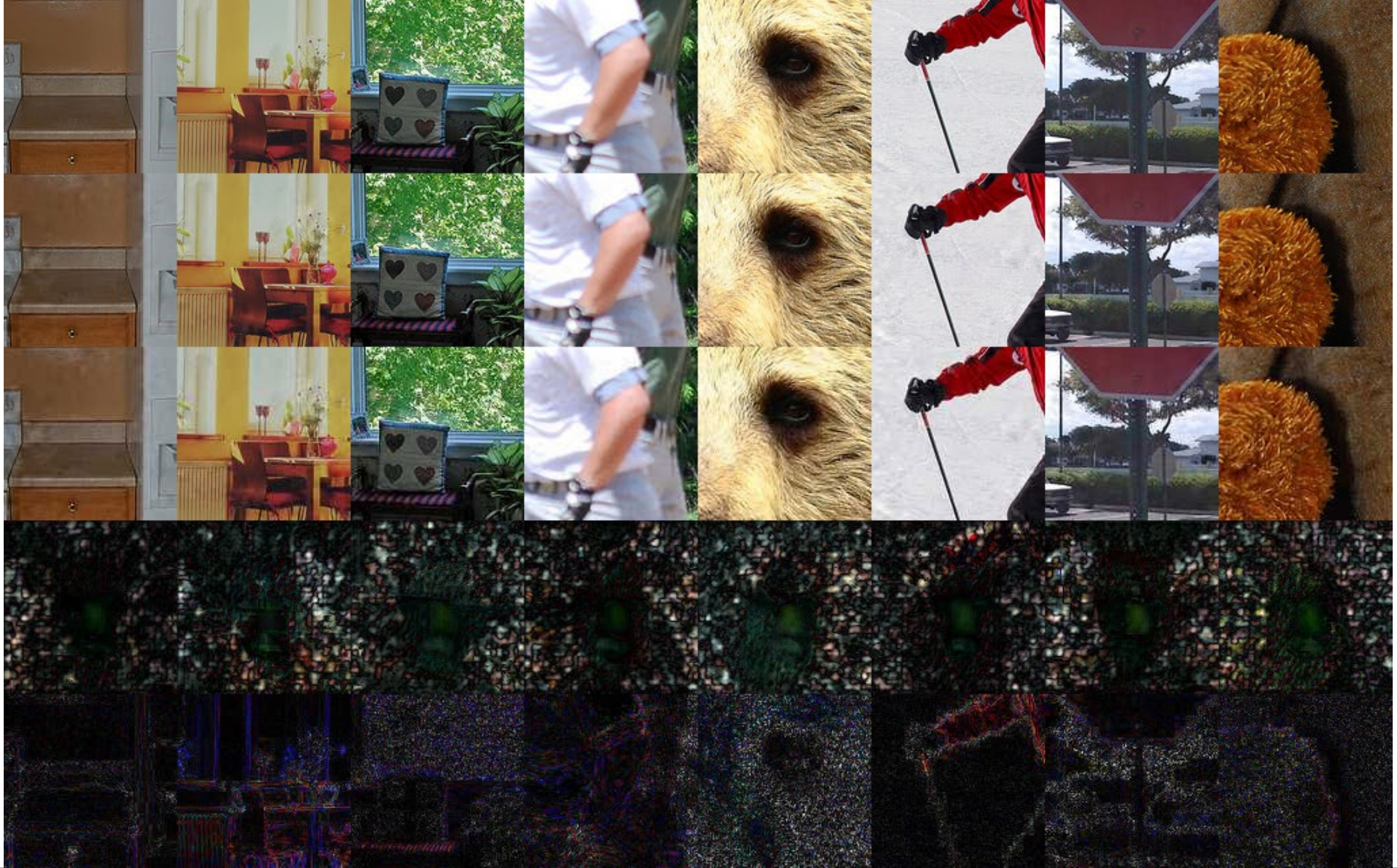}
	\caption{JpegTest. Dataset: COCO. Image size: $128\times 128$.}
	\Description{JpegTest.}
\end{figure}

\end{document}